\documentclass[10pt,twocolumn,letterpaper]{article}

\usepackage{iccv}
\usepackage{times}
\usepackage{graphicx}
\usepackage{amsmath}
\usepackage{amssymb}
\usepackage{balance}
\usepackage{pdfpages}

\usepackage{kotex}
\usepackage{comment}
\usepackage{multirow}
\usepackage[table]{xcolor}
\newcommand{\Eq}[1]  {Eq.\ (\ref{equ:#1})}

\newcommand{\Fig}[1] {Figure \ref{fig:#1}}

\newcommand{\Tbl}[1]  {Table \ref{tbl:#1}}

\newcommand{\Sec}[1] {Section \ref{sec:#1}}

\newcommand{\Etal}   {et al.}

\ificcvfinal\fi
\newcommand\blfootnote[1]{%
  \begingroup
  \renewcommand\thefootnote{}\footnote{#1}%
  \addtocounter{footnote}{-1}%
  \endgroup
}

\usepackage[pagebackref=true,breaklinks=true,letterpaper=true,colorlinks,bookmarks=false]{hyperref}

\iccvfinalcopy 

\def\iccvPaperID{****} 

\ificcvfinal\fi

\begin{document}

\title{Deep Virtual Markers for Articulated 3D Shapes}

\author{Hyomin Kim \hspace{0.5cm}
Jungeon Kim \hspace{0.5cm}
Jaewon Kam \hspace{0.5cm}
Jaesik Park$^*$ \hspace{0.5cm}
Seungyong Lee$^*$
\vspace{2.5mm} \\
POSTECH
}

\maketitle
\blfootnote{$^*$Joint corresponding authors.}
\begin{abstract}
We propose deep virtual markers, a framework for estimating dense and accurate positional information for various types of 3D data.
We design a concept and construct a framework that maps 3D points of 3D articulated models, like humans, into virtual marker labels. To realize the framework, we adopt a sparse convolutional neural network and classify 3D points of an articulated model into virtual marker labels. We propose to use soft labels for the classifier to learn rich and dense interclass relationships based on geodesic distance. To measure the localization accuracy of the virtual markers, we test FAUST challenge, and our result outperforms the state-of-the-art. We also observe outstanding performance on the generalizability test, unseen data evaluation, and different 3D data types (meshes and depth maps). We show additional applications using the estimated virtual markers, such as non-rigid registration, texture transfer, and realtime dense marker prediction from depth maps.
  
\end{abstract}
\section{Introduction}

\begin{figure}[t]
    \begin{center}
    \includegraphics[width=0.995\linewidth,page=1]{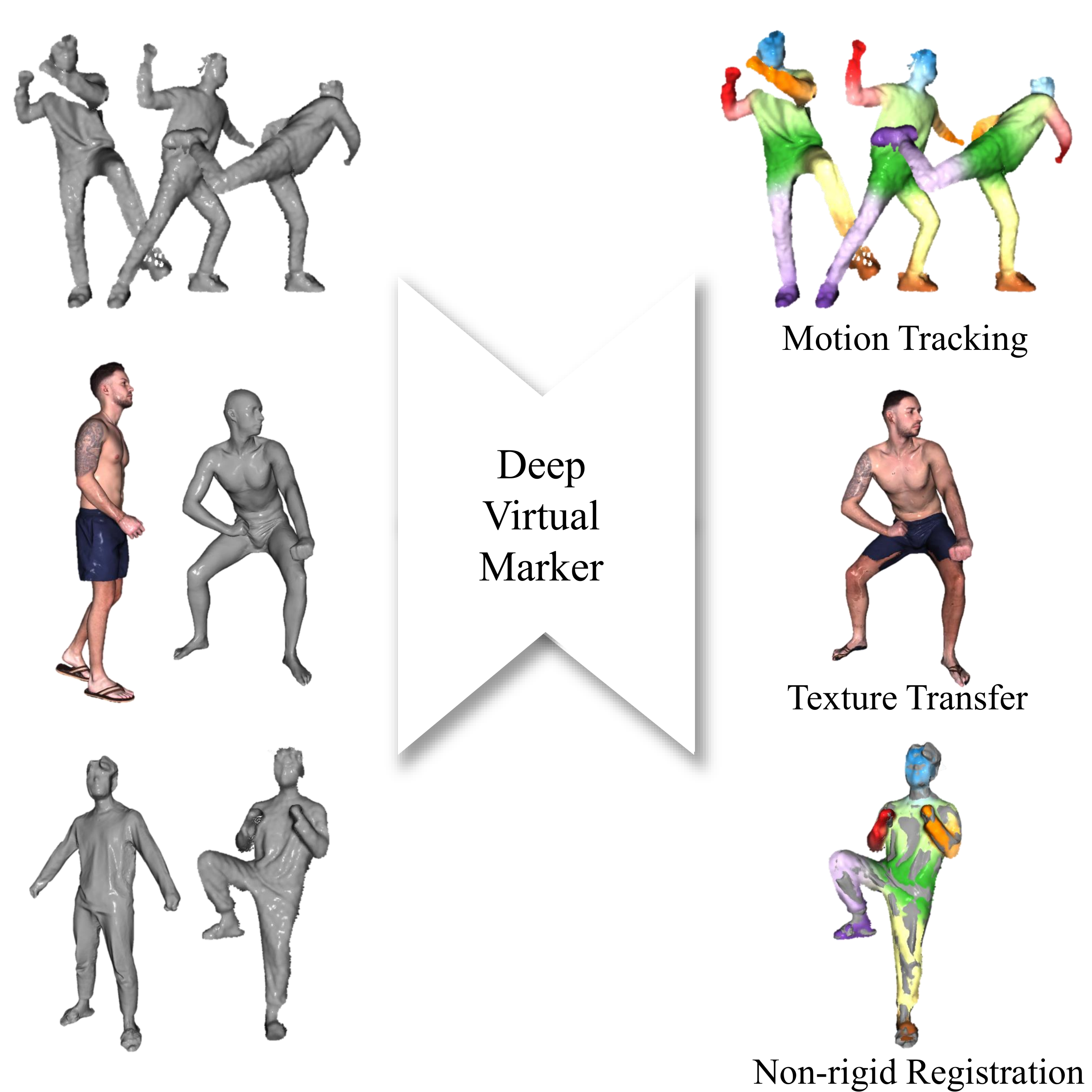}
    \end{center}
    \caption{Deep virtual markers instantly produce dense and accurate positional annotations from a mesh or depth map. Applications of deep virtual markers include motion analysis/tracking, texture transfer, and non-rigid registration.}
    \label{fig:teaser}
\end{figure}

Estimating positional information of surface points on articulated objects (such as humans and animals) is crucial for various applications such as virtual/augmented reality, movie industries, and entertainments~\cite{menache2000understanding, bregler2007motion, MOESLUND2001231,chan2010virtual}. One common way to capture such information is that actors wear special suits with beacons that are tracked by a special optical system equipped with multiple cameras. Markerless approaches~\cite{habermann2019livecap, PhysCap2020, habermann2020deepcap} have also been developed, but they are not as robust as using markers and more error-prone in difficult situations like fast motions.

We introduce \emph{deep virtual markers}, a framework that can instantly produce dense and accurate positional annotations for a 3D observation of an articulated object, such as a depth map of the object captured by a depth camera or a 3D point set sampled from the object.
In contrast to prior works that predict 3D skeletons~\cite{newell2016stacked,xia2017joint, varol2017learning} or pose parameters of parametric models~\cite{bogo2016keep, Joo_2018_CVPR}, our approach directly maps any observed 3D points to the canonical locations, providing much richer information. As a result, our approach predicts dense and reliable makers that can be a useful tool for various applications, such as 3D motion analysis, texture transfer, and non-rigid registration of human scans. Examples of the applications are shown in~\Fig{teaser}.

Our approach for dense marker prediction is learning-based, and it is built upon 3D sparse convolution~\cite{choy20194d}. Due to the fully convolutional nature of the neural network, the shape can be understood well in the cluttered or occluded cases. The proposed framework accurately determines dense marker positions via single feed-forward pass of the neural network. Therefore, our approach runs in real-time and it does not involve heuristic modules.

To train the network, we propose a new dataset that consists of realistic depth renderings of template models. 
To annotate dense supervisory labels, we begin with sparse marker annotations on the template models, and solve for heat equilibrium to obtain dense annotations, where the annotation of a surface point is represented using the relative influences from sparsely annotated points.
The annotated models are augmented with realistic motions from Mixamo~\cite{mixamo} and random poses that follows physical ranges of body joints.

The positional accuracy of the predicted virtual markers are evaluated with several benchmarks that provides precise correspondences, such as
FAUST~\cite{FAUST} and SCAPE~\cite{SCAPE}.
Computing virtual markers of the two scans is equivalent to building dense correspondences, as virtual markers provide positional annotations on common canonical domain. In the evaluation, our approach achieves the state-of-the-art performance in the non-rigid human correspondence identification benchmark. We also perform cross-validation with different datasets.

In contrast to prior works~\cite{litany2017deep, groueix20183d, prokudin2019efficient, halimi2019unsupervised, li2019lbs, deprelle2019learning, marin2020farm, ginzburg2020cyclic, gao2020learning}, our deep virtual markers are applicable to various 3D data types, such as full 3D meshes, point clouds, and depth maps. Our approach is also intended to be applied for any articulated objects. Therefore, we prepare a new data set for animals as well, and we show that our approach can handle depth map of a real cat.
Codes are publicly available.\footnote{\href{https://github.com/T2Kim/DeepVirtualMarkers}{https://github.com/T2Kim/DeepVirtualMarkers}.}

In summary, our key contributions are as follows:
\begin{itemize}
\item We propose a real-time approach to extract dense markers from 3D shapes regardless of data types, such as full 3D meshes and depth maps.
\item We propose an effective approach to annotate dense markers for template models for preparing training sets. We also propose a new dataset of dense markers.
\item Our approach achieves state-of-the-art performance on non-rigid correspondence identification task that requires accurate marker localization.
\item Experiment results show the generalizability of our approach on various datasets and data types. We also demonstrate that our approach can handle various articulated objects such as humans and cats.
\end{itemize}

\begin{table}[t]
    \centering
    \caption{Comparison with related work. Our approach is a learning-based real-time method that assigns dense markers. The proposed method can handle a mesh, partial observation (such as depth map), and multiple instances. Our approach does not require heavy pre-processing.
    In the table, FC Res., GCNN, MLP, AE, and RF indicate fully connected residual net, graph convolutional neural network, multi-layer perceptron, autoencoder, and random forest, respectively. Please see Sec.~\ref{sec:related} for the details.}
    \resizebox{.995\linewidth}{!}{
    \begin{tabular}{|c|c|c|c|c|c|c|c|c|c|c|}
        \hline
         Method & Network & 
         {\rotatebox[origin=c]{90}{Learning-based}} & 
         {\rotatebox[origin=c]{90}{Dense prediction}} & 
         {\rotatebox[origin=c]{90}{Handle mesh}} & 
         {\rotatebox[origin=c]{90}{~Handle partial obs.~}} & 
         {\rotatebox[origin=c]{90}{No topology}} & 
         {\rotatebox[origin=c]{90}{Multiple objects}} & 
         {\rotatebox[origin=c]{90}{Real-time}} & 
         {\rotatebox[origin=c]{90}{~No preprocessing~}} & 
         {\rotatebox[origin=c]{90}{Publication}}\\
         \hline
         \hline
         SmoothShells~\cite{eisenberger2020smooth} & - &&\checkmark &\checkmark &  &  &  & &&`20\\
         FARM~\cite{marin2020farm}& - && \checkmark&\checkmark &  &  &  && & `20\\
         \hline
         UnsupFMNet~\cite{halimi2019unsupervised}& AE &U& \checkmark& \checkmark &  &  &&  & &`19\\
         SurFMNet~\cite{roufosse2019unsupervised}& FC Res. &U&\checkmark& \checkmark &  &  &  &  & &`19\\
         CyclicFM~\cite{ginzburg2020cyclic} & FC Res. &Self&\checkmark& \checkmark &  &  &  &  & &`20\\
         Deep Shells~\cite{eisenberger2020deep} & GCNN &U&\checkmark& \checkmark &  &  &&  & &`20\\
         \hline
         Shotton~\Etal~\cite{shotton2011real} & RF & S& & & \checkmark & \checkmark & \checkmark &\checkmark &\checkmark&`11\\
         GCNN~\cite{masci2015geodesic}& Geo. CNN &S &\checkmark& \checkmark &  & &  &  &&`15\\
         DHBC~\cite{wei2016dense}& 2D CNN &S&\checkmark& \checkmark & \checkmark & \checkmark &  & \checkmark&\checkmark &`16\\
         Nishi~\Etal~\cite{NISHI2017402}& 2D CNN&S & & & \checkmark & \checkmark &  & &\checkmark&`17\\
         FMNet~\cite{litany2017deep}& FC Res. &S&\checkmark& \checkmark &  &  &  &   &&`17\\
         Fey~\Etal~\cite{Fey_2018_CVPR}& SplineCNN &S&\checkmark&\checkmark &  &  &  &   &\checkmark&`18\\
         FeaStNet~\cite{Verma_2018_CVPR}& GCNN &S&\checkmark& \checkmark& & & & &\checkmark&`18\\
         Lim~\Etal~\cite{lim2018simple}& FC RNN &S&\checkmark& \checkmark& & & & &&`18\\
         3D-CODED~\cite{groueix20183d}& AE &S/U&\checkmark & \checkmark &  &  &  && \checkmark&`18\\
         SpiralNet++~\cite{Gong_2019_ICCV}& SpiralCNN &S& \checkmark&\checkmark  &  &  &  && \checkmark&`19\\
         MGCN~\cite{wang2020mgcn} & GCNN &S&\checkmark& \checkmark &  &  &  && & `20\\
         GeoFMNet~\cite{donati2020deep} & 3D CNN &S&\checkmark& \checkmark &  &  &  && & `20\\
         LSA-Conv~\cite{gao2020learning}& GCN &S&\checkmark& \checkmark &  &  &  && \checkmark& `20\\
         Marin~\Etal~\cite{marin2020correspondence}& MLP&S &\checkmark& \checkmark & \checkmark &  &  && \checkmark& `20\\
         \hline
         \hline
         \rowcolor{yellow}
         Ours-oneshot & Sparse &  & \checkmark & \checkmark & \checkmark & \checkmark & \checkmark & \checkmark & \checkmark & \\
         \rowcolor{yellow}
         Ours-multiview & 3D CNN & \multirow{-2}{*}{S} & \checkmark & \checkmark & \checkmark & \checkmark & \checkmark & \checkmark &  & \multirow{-2}{*}{-}\\
         \hline
    \end{tabular}
    }
    \label{tab:overall comparison}
\end{table}

\section{Related Work}
\label{sec:related}
The proposed deep virtual marker is closely related to the prior arts that are built for body part recognition, geometric feature descriptor, and non-rigid human body alignment. We review representative approaches here, and the summary is shown in~Table~\ref{tab:overall comparison}.

\vspace{2mm}
\noindent\textbf{Body part recognition.}
Several works have been proposed to segment body parts of articulated objects using color images or depth images. Many approaches~\cite{wang2015semantic, oliveira2016deep,lassner2017unite, xia2017joint,  varol2017learning, mu2020learning, shotton2011real, hernandez2012graph, NISHI2017402} focus on human part segmentation, whereas others target for animal part segmentation~\cite{wang2015semantic,mu2020learning}. As human part segmentation and human pose estimation are closely related to each other, some works try to tackle both problems simultaneously~\cite{ xia2017joint, varol2017learning, mu2020learning}. 
Since these works usually adopt data-driven approaches, they need annotated body parts for training. 

The works by Shotton~\Etal~\cite{shotton2011real} and Nishi~\Etal~\cite{NISHI2017402} are close to our approach. Their methods use depth images and learn a classifier to assign human body part labels to depth pixels. 
However, the part labels obtained by the methods are too sparse to precisely align two human models of different shapes. In contrast, our approach generates dense labels enough for human body alignment.

\vspace{2mm}
\noindent\textbf{Geometric features.}
Traditional methods have been developed for hand-crafted geometric features~\cite{johnson1999using,frome2004recognizing, rusu2009fast, salti2014shot, sun2009concise, aubry2011wave}. Recent methods use deep neural networks (DNNs), and GCNN~\cite{masci2015geodesic}, ACNN~\cite{boscaini2016learning}, and Monet~\cite{Monti_2017_CVPR} define convolutions on non-Euclidean manifolds. Other approaches adopt the graph convolutional network (GCN) to utilize the informative connectivity of input geometry~\cite{Fey_2018_CVPR, Verma_2018_CVPR, lim2018simple, Gong_2019_ICCV, wang2020mgcn, gao2020learning}. FCGF~\cite{choy2019fully} adopts 3D sparse convolutional network~\cite{choy20194d} to compute dense geometric features. 

Our approach is largely inspired by FCGF~\cite{choy2019fully}. FCGF~\cite{choy2019fully} shows high performance on partial observations, but it is crafted for rigid registration of point clouds. Our approach extends its applicability to non-rigid articulated objects, such as humans and cats.

\vspace{2mm}
\noindent\textbf{Non-rigid shape matching.} Understanding human body shape and coherent geometric description are actively researched for 3D human body matching.
Stitched Puppet~\cite{zuffi2015stitched} devises a part-based human body model and optimizes the parameters of each part to fit the target mesh. 3D-CODED~\cite{groueix20183d} uses autoencoder architecture to deform the template mesh.
LoopReg~\cite{bhatnagar2020loopreg} proposes an end-to-end framework that utilizes a parametric model. The parameters are implicitly diffused over 3D space, and the approach utilizes a parametric model to find shape correspondence. Therefore, the representation capability of the model can restrict the final accuracy.

After its introduction for shape matching, a functional map~\cite{ovsjanikov2012functional} has led many follow-up studies. FMNet~\cite{litany2017deep} extracts SHOT descriptors~\cite{salti2014shot} and transforms them through a fully connected network. 
CyclicFM~\cite{ginzburg2020cyclic} proposes a self-supervised learning method that reduces cyclic distortions. While most functional map based methods use the Laplace-Beltrami operator~\cite{sigCourseFuntionalMaps} to construct the basis, 
Marin et al.~\cite{marin2020correspondence} uses a data-driven approach by learning the probe function. GeoFMNet~\cite{donati2020deep} uses KPConv~\cite{Thomas_2019_ICCV} to enable the network to directly extract features from an input 3D shape. 
FARM~\cite{marin2020farm} exploits non-rigid ICP algorithm to refine a functional map. Smooth Shells~\cite{eisenberger2020smooth} and Deep Shells~\cite{eisenberger2020deep} simplify the shapes by reducing spectral bases, and they align the generated multi-scale shapes progressively in a coarse-to-fine manner.

Functional map-based methods generally conduct preprocessing to construct the basis and need a mesh topology. In contrast, our method does not require a mesh topology at inference time 
and can handle various 3D data types.
\section{Method}

Our framework assigns \emph{dense virtual markers} to an articulated model. A virtual marker indicates a position on a canonical 3D shape. Our approach takes a raw point cloud and produces a virtual marker for each input point. 
Our approach is learning-based, so we prepare a new dataset for the training. For this procedure, we annotate sparse markers on a canonical 3D model and densify them to cover the whole surface. 

\subsection{Virtual markers}
\label{sec:virtual_markers}
Virtual markers are positional labels that are densely distributed on the surface of a canonical 3D shape of an articulated model. A method that predicts such positional labels can help us understand the 3D geometry. For instance, 
it can be used for direct texture map transfer between two 3D scans of human body without surface parameterization and finding correspondences, as shown in~\Fig{teaser}.

Our idea for defining virtual markers is to utilize the intrinsic structure of the shape of an articulated object. For instance, humans have four limbs, one head, and identically working joints. In this manner, even if persons' appearances are not the same, we can think of coherent markers that are placed on the human body.

Similarly to ours, assigning markers on the surface of a human model and solving the label classification problem is the common procedure used in the previous works~\cite{shotton2011real,NISHI2017402}. However, the most noticeable difference is that the prior methods use hard labels of markers to indicate separate human parts, as shown in \Fig{method[virtual marker]} (c). On the other hand, our virtual markers can smoothly indicate every single point on the surface of the human model as a soft label based on geodesic distance (\Fig{method[virtual marker]} (d)).

\subsection{Annotation}
\label{sec:virtual_markers_acquisition}
\begin{figure}[t]
	\centering
	\begin{tabular}{c}
		\includegraphics[width=0.33\textwidth]{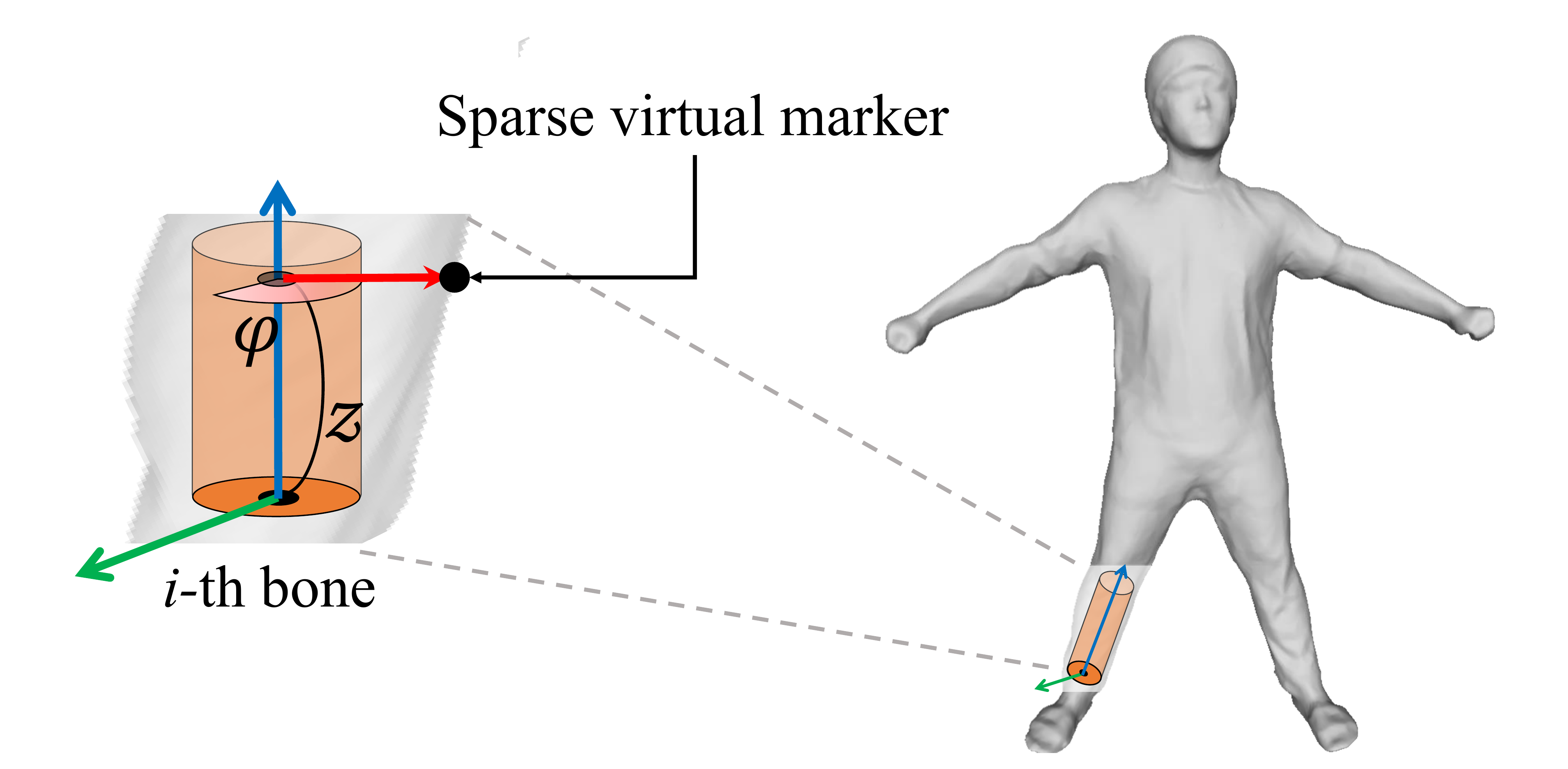} \\
		\small{(a)}\\
	\end{tabular}
	\begin{tabular}{c@{\hspace{0.1cm}}c@{\hspace{0.1cm}}c@{\hspace{0.1cm}}c}
		\includegraphics[width=0.115\textwidth]{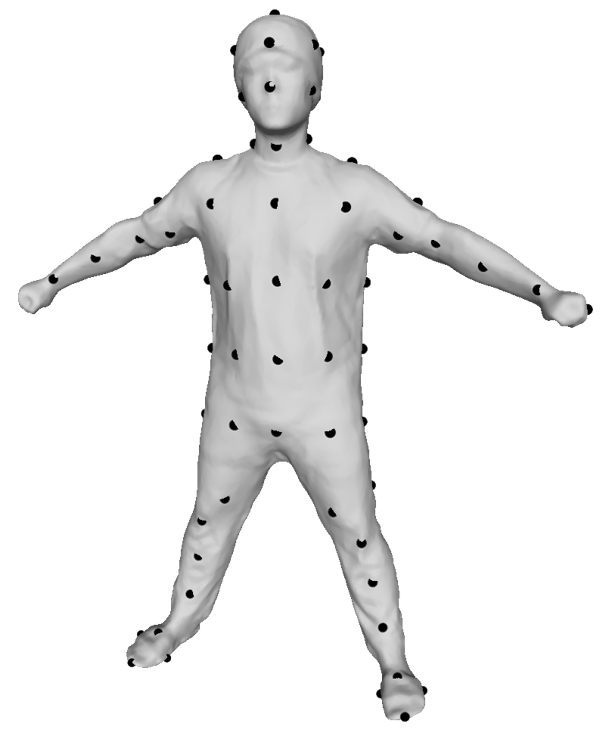} &
		\includegraphics[width=0.105\textwidth]{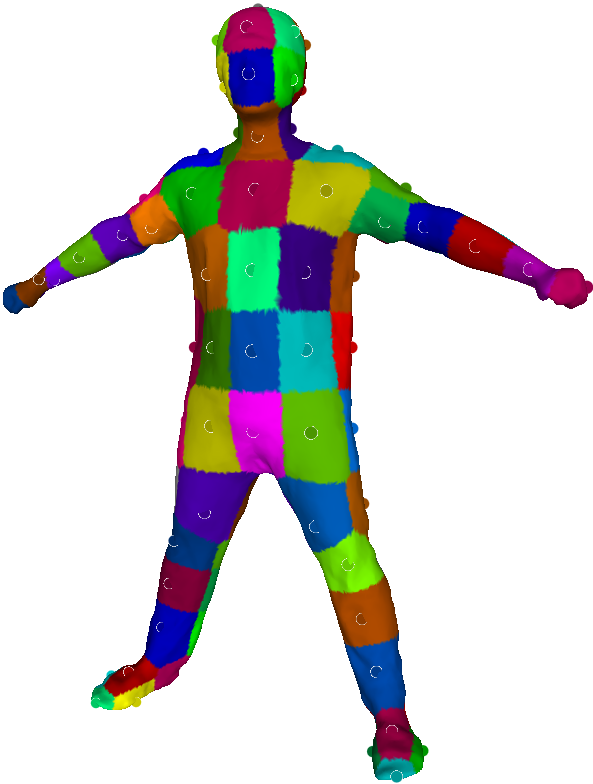} &
		\includegraphics[width=0.105\textwidth]{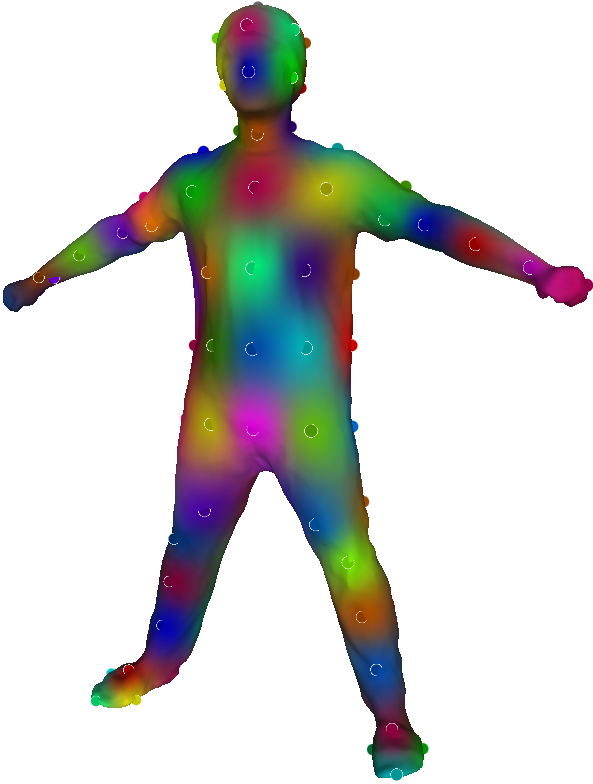} &
		\includegraphics[width=0.11\textwidth]{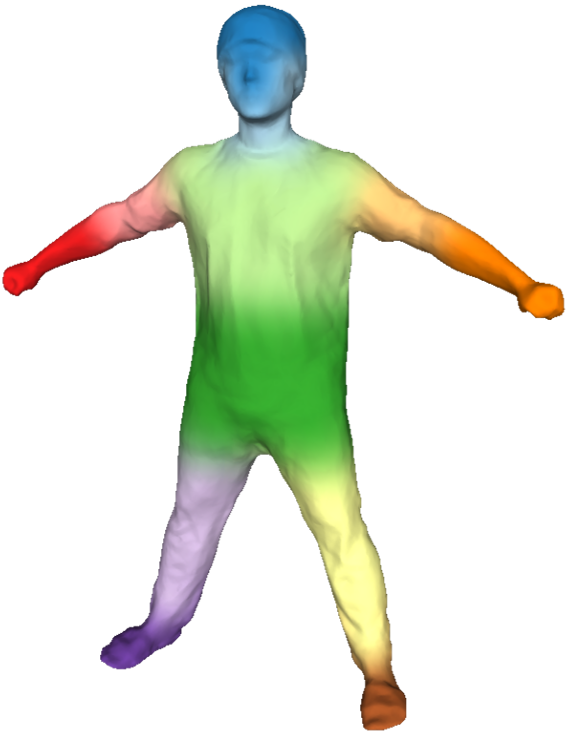}\\
		\small{(b)} & \small{(c)} & \small{(d)} & \small{(e)} \\
	\end{tabular}
	\vspace{2mm}
	\caption{Virtual markers. (a) A sparse virtual marker in the local cylindrical coordinates of a skeleton. The red arrow is the ray used for sampling the marker. (b) Sparse markers on the surface of our human model. (c) and (d) Colored human models with hard and soft labels, respectively. (e) Our color-coding scheme to visualize soft markers.}
	\label{fig:method[virtual marker]}
\end{figure}

To establish dense virtual markers for making a training dataset, we begin with annotating sparse markers on the 3D mesh model using the joint positions of a 3D human skeleton (\Fig{method[virtual marker]}~(b)). Then, the sparse markers are densely propagated to cover the whole surface of the model by solving a heat equilibrium equation (\Fig{method[virtual marker]}~(d)).
Note that this procedure is to make a dataset for training. In the test time, only a single feed-forward network is used to directly predict dense virtual markers from any type of 3D input data (depth images, point sets, or meshes).

\vspace{2mm}
\noindent\textbf{Sparse virtual markers.}
We annotate several points of interest on a mesh model with a rest pose (such as T-pose). Using the annotated points, we construct the skeleton of the template mesh model, and define sparse virtual markers. 

Consider a local cylindrical coordinate system $(\rho, \varphi, z)$ that lies on the $i$-th bone of the skeleton (\Fig{method[virtual marker]}~(a)). Origin of the coordinate system is an endpoint of the bone and its longitudinal axis corresponds to the direction of the bone. Consider rays emitted perpendicularly from the longitudinal axis, which can be specified using two coordinates $(\varphi_i, z_i)$. We uniformly sample a set of rays using the coordinates, and then the intersections of the sampled rays with the template mesh model become the sparse virtual markers from the $i$-th bone.
As the surface area of the part associated with each bone differs, we alter the number of samples for each bone. As a result, 99 markers are defined for our human model (\Fig{method[virtual marker]}~(b)). In the case of a cat model, we define 57 sparse markers using the same approach.
The detailed positions of the sparse makers are provided in the supplementary material. 

For acquiring such markers, it could be possible to use automatic rigging methods to detect the skeleton of the model~\cite{baran2007automatic, xu2020rignet}. However, we experimentally found that these automatic methods often fail for challenging shapes. Instead, we hired experienced 3D annotators to assign a precise bone structure by user interaction. The user annotations include joint positions and bone orientations (polar axes of local cylindrical coordinate systems). 

\vspace{2mm}
\noindent\textbf{Dense virtual markers.}
Given sparse markers, we may classify the surface points by assigning one-hot vectors (or hard labels) to the surface points. This process would result in a conventional part segmentation~\cite{shotton2011real, NISHI2017402} dataset, as shown in~\Fig{method[virtual marker]}~(c). However, such regional labels would not consider any fine-level relationships with multiple sparse markers.

In this work, instead, we use a \emph{soft} label scheme. Given $S$ sparse markers, the label for an arbitrary surface point is represented as a $S$-dimensional vector, whose $i$-th element indicates an affinity with the $i$-th sparse marker. That is, a soft label can have multiple non-zero entries, differently from a hard label. 

We acquire a soft label for every point on the surface by solving for heat equilibrium~\cite{baran2007automatic}.
Let $\mathcal{S}=\{s|1\leq s \leq S\}$ denote the set of indices for sparse markers, and $\mathcal{D}=\{d|1\leq d \leq D\}$ be the set of surface point indices on the template mesh.
We define $\boldsymbol{L}\in \mathbb{R}^{S\times D}$, where $l_{s,d}$ indicates affinity between the $d$-th point on the mesh with the $s$-th sparse marker. A column vector of $\boldsymbol{L}$ is normalized, so that $\sum_s {l}_{s,d}=1$.
If we define $\boldsymbol{L}^\top=\boldsymbol{W}$ that satisfy ${l}_{s,d}:=w_{d,s}$, the weight vector $\boldsymbol{w}_s=[w_{1,s}, w_{2,s}, ... ,w_{D,s}]^\mathsf{T}$ of the sparse markers $s$ for each surface point is calculated by solving the following sparse linear system:
\begin{equation}
\label{equ:heat_eq}
    -\Delta \boldsymbol{w}_s + \boldsymbol{H} \boldsymbol{w}_s = \boldsymbol{H} \boldsymbol{p}_s,
\end{equation}
where $-\Delta$ is the Laplacian operator of the mesh, $\boldsymbol{p}_s$ is a binary indicator vector, $\boldsymbol{H}\in \mathbb{R}^{D\times D}$ is the diagonal matrix having the heat contribution of the sparse marker $s$ to vertex $d$\footnote{We denote the $s$-th sparse marker as $s$, and the $d$-th vertex as $d$ for notational convenience.}. 
If sparse marker $s$ is nearest to vertex $d$, $p_{s,d}$ is set to $1$, and
the $d$-th diagonal component $H_{d,d}$ of matrix $\boldsymbol{H}$ is defined as $\frac{c}{\mathcal{D}(d,s)^2}$, where $c$ is a constant (we use $c=1$) and $\mathcal{D}(d,s)$ is the geodesic distance from vertex $d$ to sparse marker $s$. If $k$ virtual markers are equidistant from vertex $d$, $p_{s,d}=\frac{1}{k}$ and $H_{d,d}=\frac{kc}{\mathcal{D}(d,s)^2}$.
Otherwise, $p_{s,d}$ becomes $0$.

We solve the heat equilibrium condition in~\Eq{heat_eq} for every sparse marker. As a result, we have the weight matrix $\boldsymbol{W}=[\boldsymbol{w}_1 , \boldsymbol{w}_2 , ... , \boldsymbol{w}_S]$. The $d$-th row of matrix $\boldsymbol{W}$ becomes the weights of all sparse markers for vertex $d$. We set the $d$-th row of matrix $\boldsymbol{W}$ to the soft label $\boldsymbol{l}_{:,d}$ for vertex $d$. The computed soft labels are visualized in~\Fig{method[virtual marker]}~(e).

This approach is related to the work of Baran~\etal~\cite{baran2007automatic}. They compute weights between vertices and skeleton for mesh skinning. We extend the idea for computing soft labels for vertices on the template mesh. Besides, instead of Euclidean distance used in~\cite{baran2007automatic}, we use the \emph{geodesic distance} between a vertex $d$ and a soft marker $s$ to acquire ${l}_{s,d}$. Computing soft labels takes about 150 seconds per model, but needs to be done only for training data preparation. 

\subsection{Dataset}
\label{sec:dataset}
With the proposed approaches to define sparse and dense markers, we build two datasets to train a neural network.

\vspace{2mm}
\noindent\textbf{Human dataset.}
The dataset is made with 33 full mesh models of humans. They are captured with Doublefusion~\cite{yu2018doublefusion} (6 models) and collected from Renderpeople~\cite{renderpeople} (17 models) and FAUST~\cite{FAUST} (10 models). We subsample meshes of the human models to have 60k vertices for the annotation. The manual annotation is conducted by experienced annotators to provide 33 consistent joint positions. 

To deal with various non-rigid deformations, we augment each template mesh model with diverse motions. We obtained 15,000 motions from Mixamo~\cite{mixamo}. Note that we \emph{deform the annotated meshes}, so we do not need to annotate the deformed models again. We use linear blend skinning (LBS)~\cite{kavan2007skinning} to animate the template models.
In addition to this pose augmentation, we deploy random movements to the joints of the body. The random motions are bounded by physical ranges of the joints. In this manner, we can obtain a diverse dataset with different people and poses.

In addition to the pose augmentation, we extend the dataset for partial geometric observations, such as depth maps. We produce synthetic depth maps by rendering densely annotated mesh models from arbitrary viewpoints. Since a rendered depth pixel knows its original position in the annotated mesh, each depth pixel obtains a proper soft label. We used 40 viewpoints for rendering.

Note that we augment the densely annotated mesh model using the Mixamo and random poses \emph{on-the-fly} in the training phase. For the case of depth map generation, the dataset is augmented even more by using random viewpoints. This scheme introduces great diversity in our training dataset.

\vspace{2mm}
\noindent\textbf{Cat dataset.}
To demonstrate that our approach is applicable to other classes of articulated objects, we prepare another dataset for an animal shape. We select two cat models from SMAL~\cite{zuffi20173d}, and use the same procedure for dense annotation of the models.

\subsection{Training}
We employ the ResUNet architecture based on sparse tensors~\cite{choy20194d} to classify 3D points of an articulated model. Details of the network architecture is provided in the supplementary material. The classifier directly utilizes soft labels desribed in~\Sec{virtual_markers_acquisition} for the supervision.

Our network employs a multi-class cross entropy loss defined as follows:
\begin{equation}
    \mathcal{L} = -\sum_d^D\sum_s^S{l_{s,d} \log{(\frac{l'_{s,d}}{\sum_j^S{e^{l'_{j,d}}}})}},
    \label{equ:MLCE}
\end{equation}
where ${l}_{s,d}$ denotes the $s$-th element of a soft label $\boldsymbol{l}_{:,d}$, and ${l'_{s,d}}$ is the $s$-th element of the inferred soft label ${\boldsymbol{l}'_{:,d}}$ for a vertex $d$. $D$ are the number of vertices, and $S$ is the number of sparse markers. 

As shown in~\Fig{method[virtual marker]}~(e), soft labels encode the smooth geometric relationships with adjacent surface points on the template mesh model. Consequently, without using any regularization on the smoothness of the prediction, our trained network gets to predict \emph{smoothly varying} virtual markers.

\vspace{2mm}
\noindent\textbf{Two approaches.}
We propose to train and test our network with two different configurations.

(1) \textbf{\emph{Ours-oneshot.}}
In the first setting, for training the network, we use annotations for full mesh models together with partial observations (depths). In this case, the network can infer dense virtual markers for any input data (partial or full) instantly without per-view prediction and merging. This approach saves substantial computation time when inferring virtual markers for full mesh models.

(2) \textbf{\emph{Ours-multiview.}} In the second setting, we use only partial observations (3D point cloud of depth maps) to train the network. Then, to handle a full 3D model, we use the rendered depths of the model from 72 different viewpoints. We feed each rendered depth to our network and obtain dense virtual markers for 72 views. Finally, we aggregate them to obtain the dense virtual markers for the whole 3D model. For the aggregation, we convert and combine the 72 rendered depth maps to a single point set, and the virtual marker for a vertex of the 3D model is determined by weighted averaging the virtual markers of the $k$-nearest neighbors in the point set. This approach is slower than \emph{Ours-oneshot}, but it produces more reliable reasult due to the multiview concensus.

\subsection{Timing complexity} 
We train and test our network with a workstation equipped with an Intel i7-7700K 4.2GHz CPU, 64GB RAM, and NVIDIA Titan RTX, Quadro 8000 GPU.
The number of network parameters is about 38 million. Training takes about $3 \sim 4$ and $8 \sim 10$ seconds per iteration using partial observations (depths) and full mesh models, respectively. 
In test time, inference time is about $0.05 \sim 0.07$ seconds per a depth map or 3D point set, nearly achieving 20 frames per second.

\begin{figure}[t]
	\centering
	\begin{tabular}{c@{\hspace{0.3cm}}c@{\hspace{0.1cm}}c}
		\includegraphics[width=0.1\textwidth]{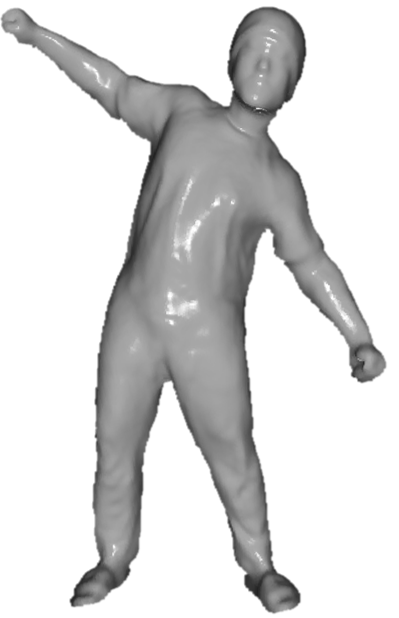} &
		\includegraphics[width=0.12\textwidth]{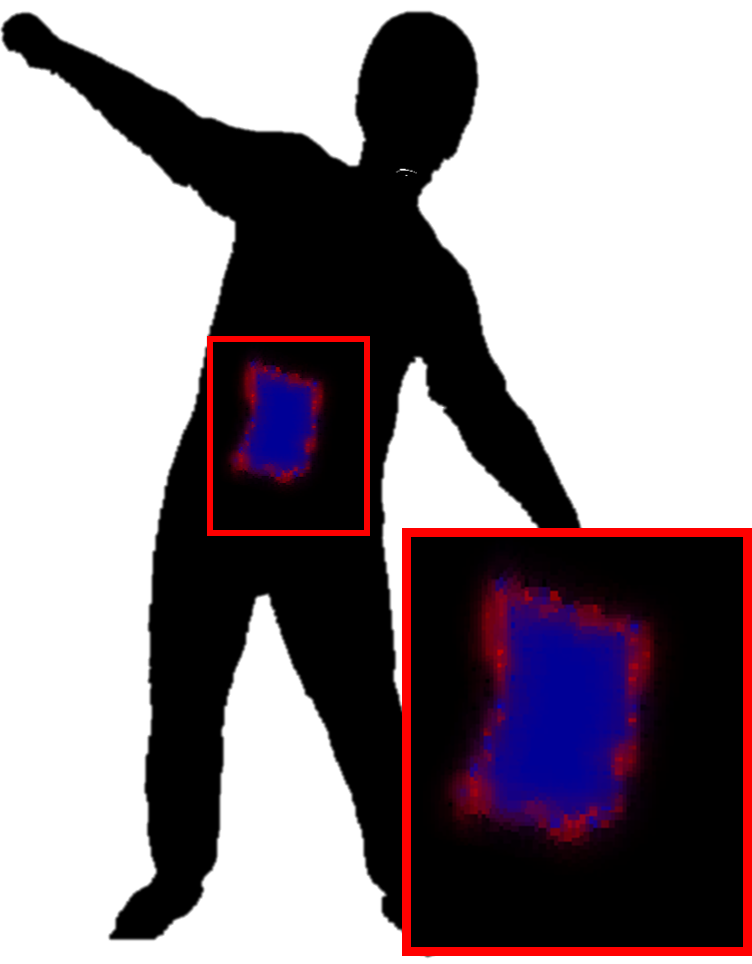} &
		\includegraphics[width=0.12\textwidth]{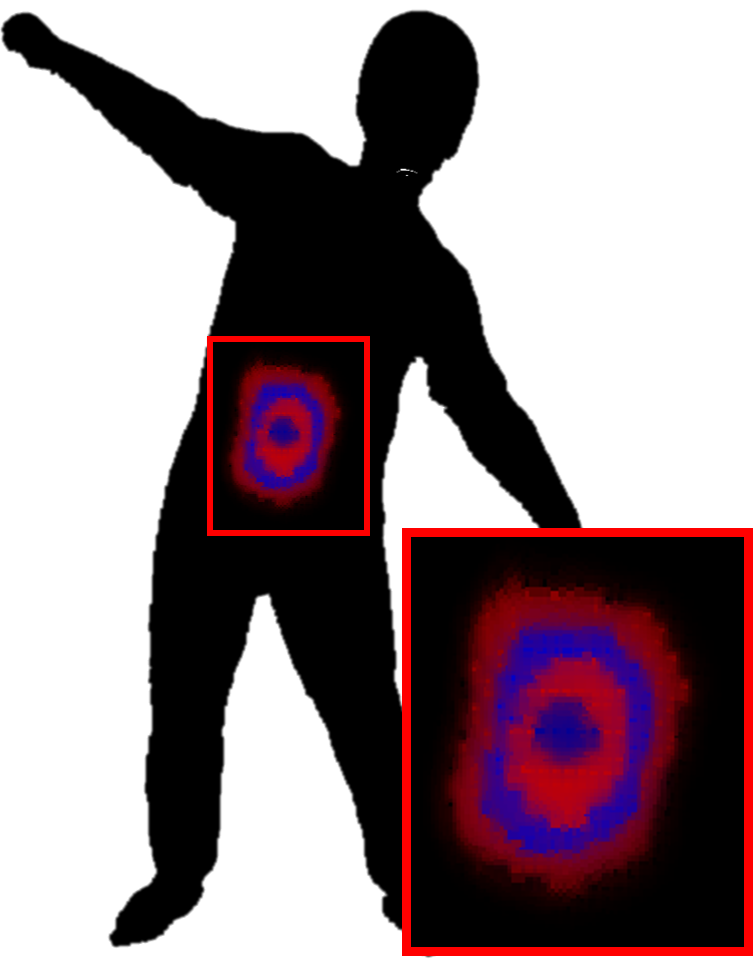}\\
		\small{(a)} & \small{(b)} & \small{(c)} \\
	\end{tabular}
	\caption{Effects of soft labels. (a) A reference 3D model. (b) and (c) Visualizations of the weights of a sparse marker obtained from networks trained with hard (one-hot) labels and soft labels as supervision, respectively. Red and blue stripes show alternating weight intervals with gap of 0.2 in the range of [0.2 to 1]. Soft labels induce smooth isotropic weight distribution.}
	\vspace*{-6pt}
	\label{fig:app[vshard]}
\end{figure}

\section{Results}
\subsection{Effectiveness of soft labels}
To analyze the effects of soft labels, we trained two networks using hard labels and soft labels. \Fig{app[vshard]} visualizes the weights of a sparse marker obtained using the two networks. We can find that the network trained with soft labels outputs smoothly varying virtual markers.

\subsection{Shape correspondence}
To validate the accuracy of virtual markers, we extensively evaluate with shape correspondence challenges. The motivations for this experiment are as follows. (1) A large-scale public dataset to measure the accuracy of dense 3D markers is not present. (2) Dataset for finding shape correspondences is well-established instead, and such datasets are captured with high-end 3D capturing systems. (3) Good performance on finding correspondences indicates consistent and coherent marker positions for diverse poses.

\vspace{2mm}

\begin{figure}[t]
	\centering
 	\setlength\tabcolsep{1pt}
 	\renewcommand{\arraystretch}{0.7}
	\begin{tabular}{c c c c c c}
 	&&&\multicolumn{3}{r}{ 
        \includegraphics[width=0.19\textwidth]{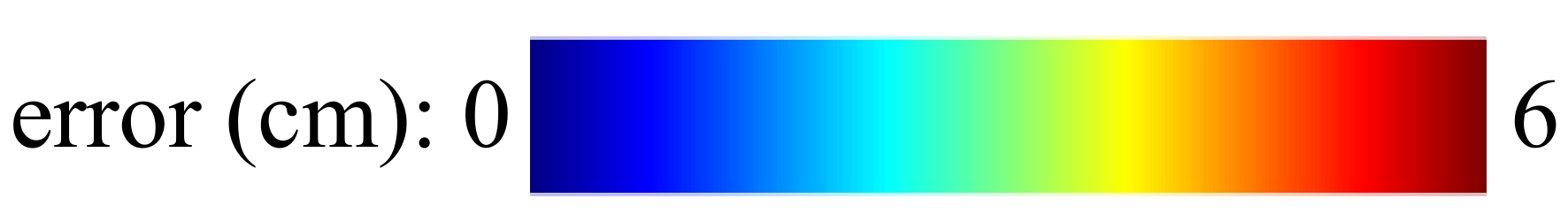}}\\
		\includegraphics[width=0.07\textwidth]{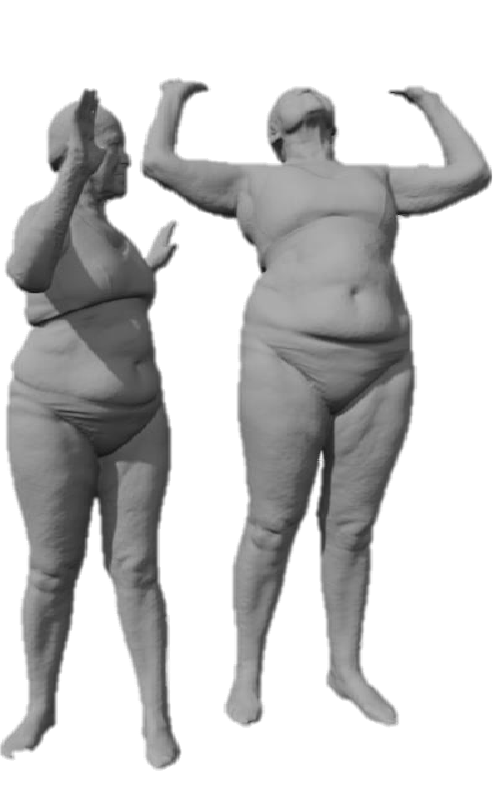} &
		\includegraphics[width=0.07\textwidth]{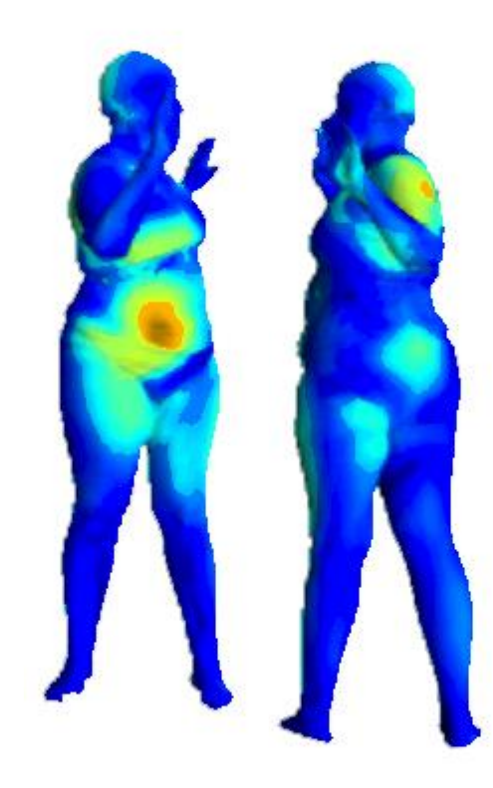} &
		\includegraphics[width=0.07\textwidth]{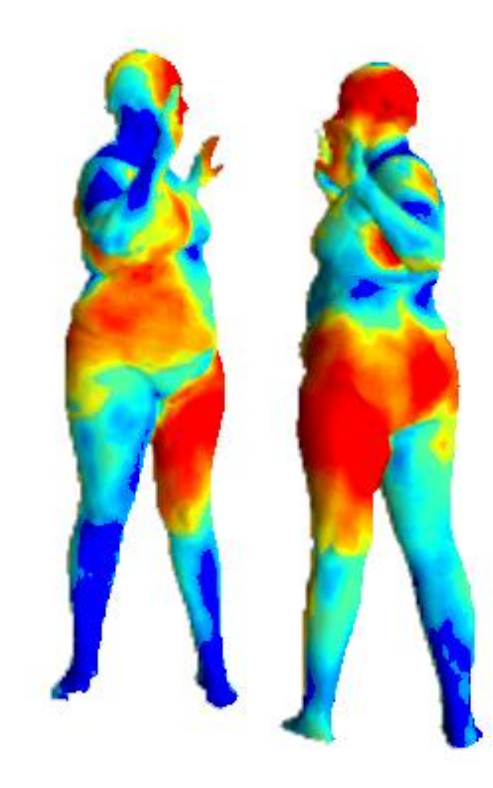} &
		\includegraphics[width=0.07\textwidth]{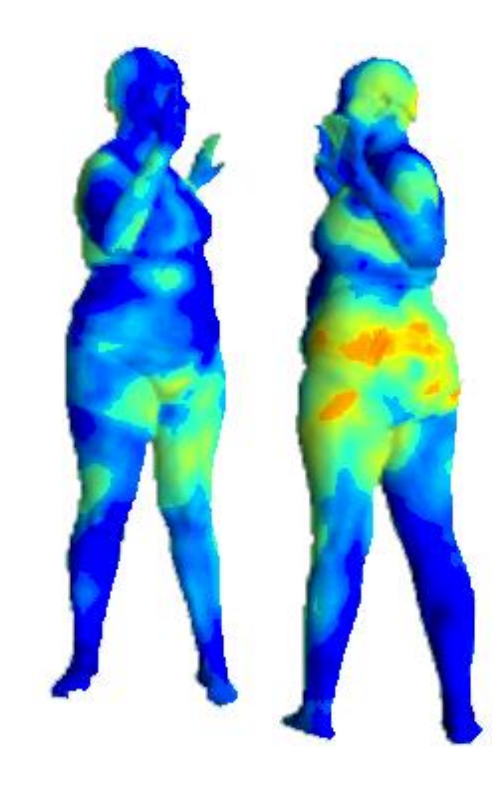} &
		\includegraphics[width=0.07\textwidth]{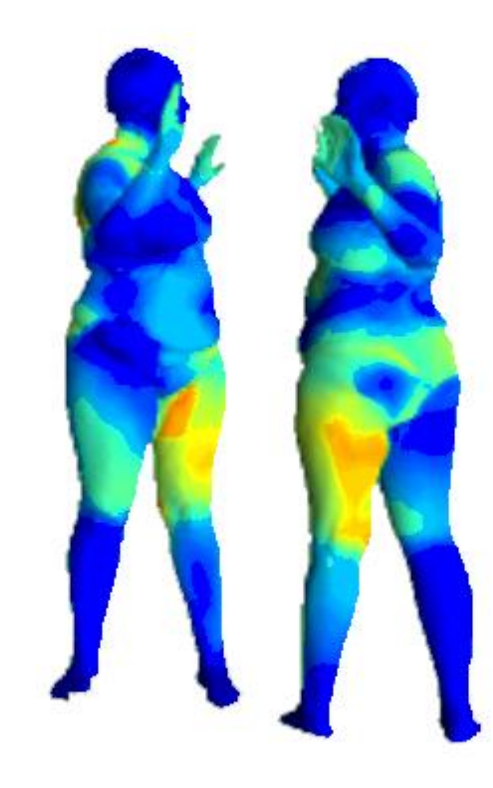} &
		\includegraphics[width=0.07\textwidth]{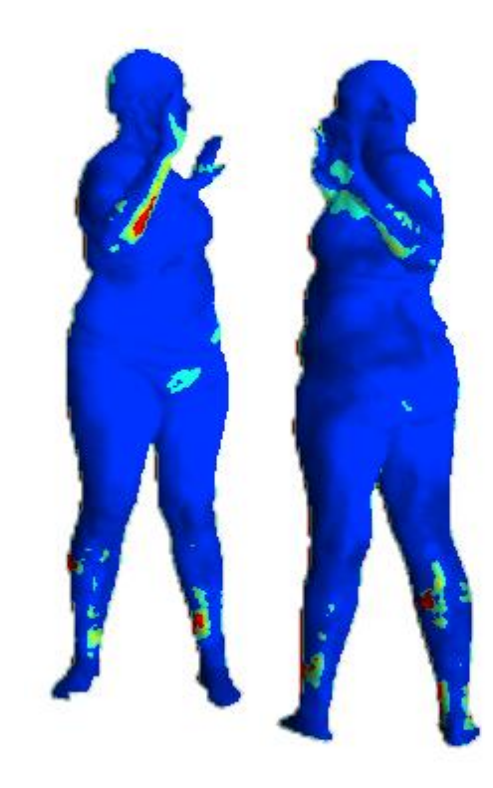} \\
        \small{(a)} &
		\small{(b)~\cite{prokudin2019efficient}} &
		\small{(c)~\cite{groueix20183d}} &
        \small{(d)~\cite{deprelle2019learning}} &
		\small{(e)~\cite{zuffi2015stitched}} & 
		\small{(f) Ours} \\
	\end{tabular}
	\caption{
	Example of FAUST correspondence finding challenge. We compare our results with BPS~\cite{prokudin2019efficient}, 3D-CODED~\cite{groueix20183d}, AtlasNetV2~\cite{deprelle2019learning}, SP~\cite{zuffi2015stitched}. The matching between two meshes on the leftmost is obtained, and its accuracy is visualized using a jet color map. This is an example of the intra-subject test set, where the two meshes are captured from the same person with different poses.
	}
	\label{fig:exp[various_comparison_FAUST_Challenge]}
	\vspace*{-6pt}
\end{figure}
\noindent\textbf{FAUST challenge benchmark} is the standard benchmark to measure the accuracy of finding correspondences from two high-quality human mesh models~\cite{FAUST}. This challenge is a good fit for validating the proposed \emph{deep virtual markers}, since the accurate markers should be localized for precise matching between non-rigid models.

The total dataset consists of 10 persons, and each person exhibits arbitrary 30 poses. Its ground-truth correspondence is not publicly available, so we submitted our results and obtained the evaluation results.
The benchmark test set provides pairs of two full 3D scans. Each pair shows extreme pose variations from the same person (called \emph{intra-subject} challenge) or different persons (called \emph{inter-subject} challenge). 
Since our approach is not restricted to the person's identity or pose variations, we report our performance for both tasks. 
The FAUST experiment is also a good way to check the generalization performance on pose and identity variations.

For comparison, we refer to the numbers in the public FAUST benchmark leaderboard. The leaderboard includes the results of approaches that require manual annotations. Hence we compare only with the methods that do not require manual inputs. The results are shown in~\Tbl{FAUST_Challenge}. 
Our approach shows the most accurate results compared with prior works on both the inter- and intra-subject challenges. Visualizations of errors are shown in \Fig{exp[various_comparison_FAUST_Challenge]}.

\vspace{2mm}
\noindent\textbf{Comparison with DHBC~\cite{wei2016dense}.}
The original FAUST challenge only tackles the correspondence for two full meshes, which is called full-to-full correspondence identification. However, handling \emph{partial} observations, such as depth maps, is also important for practical applications. For this reason, we additionally compare our approach with DHBC~\cite{wei2016dense} that can handle full-to-full, full-to-partial, and partial-to-partial dense correspondences. 

Since DHBC is not shown on the FAUST leaderboard~\cite{FAUST}, we conduct the comparison by following the evaluation protocol suggested by Chen~\Etal~\cite{chen2015robust}. The protocol uses some pairs of 3D scans in the FAUST training set and performs inter- and intra-subject challenges. 
The protocol reports the average error on all pairs (AE) and the average error on the worst pair (WE). 
\Tbl{FAUST_DHBC} shows the results of the experiment on the full-to-full case. We also added the results from RobuxtCovex~\cite{chen2015robust}. Our approach favorably outperforms all the cases. In particular, our results show about three times smaller intra-subject and inter-subject WEs than other baselines. Besides, in our case, WEs are similar to AEs. This demonstrates that our approach is reliable and robust to different persons' identities and various poses.

In addition to the full-to-full comparison, we also perform a quantitative comparison with DHBC~\cite{wei2016dense} with \emph{full-to-partial} and \emph{partial-to-partial} 3D model pairs. The new pairs are obtained by rendering 3D scans from specific viewpoints, and we used FAUST~\cite{FAUST} and SCAPE~\cite{SCAPE} datasets for the rendering. The results are shown in \Tbl{FAUST_DHBC_F2P_P2P}. Our method outperforms DHBC~\cite{wei2016dense} in these cases too.

\begin{table}[t]
    \centering
    \caption{FAUST challenge results (error in cm). We omit the results of approaches that require manual annotations in the test time. Approaches are classified into learning-based (first three rows) and non-learning-based methods (the remaining rows).}
    \resizebox{.8\linewidth}{!}{
    \begin{tabular}{c|c|c}
        \hline
        & \multicolumn{2}{c}{Type of challenge}\\
        \cline{2-3}
        Method & Intra-subject & Inter-subject \\
        \hline
        SP~\cite{zuffi2015stitched} & 1.568 & 3.126 \\
        RobustCovex~\cite{chen2015robust} & 4.860 & 8.304 \\
        Smooth Shells~\cite{eisenberger2020smooth} & - & 3.929 \\
        \hline
        FMNet~\cite{litany2017deep} & 2.436 & 4.826 \\
        3D-CODED~\cite{groueix20183d} & 1.985 & 2.769 \\
        BPS~\cite{prokudin2019efficient} & 2.010 & 3.020 \\
        Unsup FMNet~\cite{halimi2019unsupervised} & 2.510 & - \\
        LBS-AE~\cite{li2019lbs} & 2.161 & 4.079 \\
        AtlasNetV2~\cite{deprelle2019learning} & 1.626 & 2.578 \\
        FARM~\cite{marin2020farm} & 2.810 & 4.123 \\
        Cyclic FM~\cite{ginzburg2020cyclic} & 2.120 & 4.068 \\
        LSA-Conv ~\cite{gao2020learning} & - & 2.501 \\
        \hline
        Ours-oneshot & \cellcolor{yellow!20}{1.417} & \cellcolor{yellow!20}{2.495} \\
        Ours-multiview & \cellcolor{yellow!100}\textbf{1.185} & \cellcolor{yellow!100}\textbf{2.372} \\
        \hline
    \end{tabular}
    }
    \label{tbl:FAUST_Challenge}
\end{table}

\begin{table}[t]
    \centering
    \caption{Additional comparison with DHBC~\cite{wei2016dense} and RobustCovex~\cite{chen2015robust} using the FAUST training dataset (error in cm).}
    \resizebox{.995\linewidth}{!}{
    \begin{tabular}{c c c c c}
        \hline
         & Intra AE & Intra WE & inter AE & inter WE\\
        \hline
        RobustCovex~\cite{chen2015robust} & 4.49 & 10.96 & 5.95 & 14.18 \\
        DHBC~\cite{wei2016dense} & \cellcolor{yellow!20}{2.00} & 9.98 & \cellcolor{yellow!20}{2.35} & 10.12 \\
        \hline
        Ours-oneshot & 2.26 & \cellcolor{yellow!20}{3.93} & 2.44 & \cellcolor{yellow!20}{4.56}\\
        Ours-multiview & \cellcolor{yellow!100}\textbf{1.93} & \cellcolor{yellow!100}\textbf{2.76} & \cellcolor{yellow!100}\textbf{2.12} & \cellcolor{yellow!100}\textbf{3.25}\\
        \hline
        
    \end{tabular}
    }
    \label{tbl:FAUST_DHBC}
\end{table}

\begin{table}[h]
    \centering
    \caption{Finding correspondences using full-to-partial (F2P) and partial-to-partial (P2P) datasets. We compare our results with DHBC~\cite{wei2016dense} using the FAUST~\cite{FAUST} and SCAPE~\cite{SCAPE} datasets for training. For each three numbers, the first one is the error of DHBC, and the second and third are the errors of ours-multiview and ours-oneshot, respectively (error in cm).} 
    \resizebox{.995\linewidth}{!}{
        \begin{tabular}{c|c|cccc}
            \hline
            \multirow{2}{*}{Data} & \multirow{2}{*}{Type} & \multicolumn{4}{c}{\begin{tabular}[c]{@{}c@{}}Method\\ (DHBC~\cite{wei2016dense}, Ours-multiview, Ours-oneshot) \end{tabular}} \\ \cline{3-6} 
             &  & \begin{tabular}[c]{@{}c@{}}Intra AE\end{tabular} & \begin{tabular}[c]{@{}c@{}}Intra WE\end{tabular} & \begin{tabular}[c]{@{}c@{}}Inter AE\end{tabular} & \begin{tabular}[c]{@{}c@{}}Inter WE\end{tabular} \\ \hline
            \multirow{6}{*}{FAUST~\cite{FAUST}} & \multirow{3}{*}{F2P} & 6.52 & 14.61 & 6.75 & 11.79 \\
             &  & \cellcolor{yellow!100}\textbf{2.39} & \cellcolor{yellow!100}\textbf{3.52} & \cellcolor{yellow!20}2.63 & \cellcolor{yellow!20}4.62 \\
             &  & \cellcolor{yellow!20}2.41 & \cellcolor{yellow!20}4.40 & \cellcolor{yellow!100} \textbf{2.58} & \cellcolor{yellow!100} \textbf{4.52} \\ \cline{2-6} 
             & \multirow{3}{*}{P2P} & 9.51 & 22.46 & 9.81 & 32.27 \\
             &  & \cellcolor{yellow!100}\textbf{3.40} & \cellcolor{yellow!100}\textbf{11.06} & \cellcolor{yellow!100}\textbf{3.45} & \cellcolor{yellow!100}\textbf{11.166} \\
             &  & \cellcolor{yellow!20}3.96 & \cellcolor{yellow!20}20.24 & \cellcolor{yellow!20}3.74 & \cellcolor{yellow!20}16.06 \\ \hline
            \multirow{6}{*}{SCAPE~\cite{SCAPE}} & \multirow{3}{*}{F2P} & 4.33 & 11.76 & - & - \\
             &  & \cellcolor{yellow!100}\textbf{3.11} & \cellcolor{yellow!100}\textbf{4.92} & - & - \\
             &  & \cellcolor{yellow!20}3.23 & \cellcolor{yellow!20}7.29 & - & - \\ \cline{2-6} 
             & \multirow{3}{*}{P2P} & 17.19 & 38.21 & - & - \\
             &  & \cellcolor{yellow!100}\textbf{8.32} & \cellcolor{yellow!100}\textbf{27.76} & - & - \\
             &  & \cellcolor{yellow!20}8.99 & \cellcolor{yellow!20}36.94 & - & - \\ \hline
        \end{tabular}
    }
    \label{tbl:FAUST_DHBC_F2P_P2P}
\end{table}

\begin{table}[th]
    \centering
    \caption{Normalized mean geodesic error (in \%) using the dataset by Ren~\Etal~\cite{ren2018continuous}. This experiment validates the cases with various training and test sets. F and S denote FAUST and SCAPE, respectively. From top to bottom, the previous methods compared in this table are classified into three categories; axiomatic (no training), unsupervised learning, and supervised learning. The numbers except for ours are courtesy of~\cite{eisenberger2020deep}.} 
    \resizebox{.88\linewidth}{!}{
    \begin{tabular}{c|r|r|r|r}
        \hline
        & \multicolumn{4}{c}{Test / Training dataset} \\
        \cline{2-5}
        Method & \multicolumn{1}{c|}{F / F} & \multicolumn{1}{c|}{S / S} & \multicolumn{1}{c|}{F / S}& \multicolumn{1}{c}{S / F} \\
        \hline
        \hline
        BCICP~\cite{ren2018continuous} & 15.0 & 16.0 & - & -\\
        ZoomOut~\cite{melzi2019zoomout} & 6.1 & 7.5 & - & -\\
        Smooth Shells~\cite{eisenberger2020smooth} & 6.1 & 7.5 & - & -\\
        \hline
        SurFMNet+ICP~\cite{roufosse2019unsupervised} & 7.4 & 6.1 & 23.0 & 19.0\\
        Unsup FMNet+pmf~\cite{halimi2019unsupervised} & 5.7 & 10.0 & 9.3 & 12.0\\
        Deep Shells~\cite{eisenberger2020deep} & \cellcolor{yellow!20}1.7 & \cellcolor{yellow!20}2.5 & \cellcolor{yellow!100}\textbf{2.7} & 5.4\\
        \hline
        FMNet+pmf~\cite{litany2017deep} & 5.9 & 6.3 & 14.0 & 11.0\\
        3D-CODED~\cite{groueix20183d} & 2.5 & 31.0 & 33.0 & 31.0\\
        GeoFMNet+zo~\cite{donati2020deep} & 1.9 & 3.0 & 4.3 & 9.2\\
        \hline
        Ours-oneshot & 2.1 & 2.7 & 4.9 & \cellcolor{yellow!20}{3.1}\\
        Ours-multiview & \cellcolor{yellow!100}\textbf{1.5} & \cellcolor{yellow!100}\textbf{2.0} & \cellcolor{yellow!20}{4.1} & \cellcolor{yellow!100}\textbf{1.9}\\
        \hline
    \end{tabular}
    }
    \label{tbl:FAUST_SCAPE_generalize}
\end{table}

\vspace{2mm}
\noindent\textbf{Various training and test sets.}
We verify the generalizability of our method by changing the training and test sets. For this experiment, we use the dataset provided by Ren~\Etal~\cite{ren2018continuous}. The dataset contains re-meshed versions of FAUST~\cite{FAUST} and SCAPE~\cite{SCAPE} datasets. 

\Tbl{FAUST_SCAPE_generalize} shows the results, where the evaluation is based on the protocol proposed by Kim~\etal~\cite{Kim2011BIM}. SCAPE dataset contains a single human model with various poses, and FAUST includes several models with various poses. Hence, for the case of testing on FAUST and training on SCAPE (F / S), the matching accuracies of some approaches~\cite{roufosse2019unsupervised,litany2017deep,groueix20183d} are much lower than other combinations since the variety of model shapes for training is insufficient. 

Note that some of the approaches~\cite{litany2017deep, groueix20183d, donati2020deep} in \Tbl{FAUST_SCAPE_generalize} use the ground-truth correspondences for training. 
In contrast, we only use skinned synthetic shapes for training and augment the training set using Mixamo poses. It is clearly beneficial to utilize available human motions because animated models are readily obtainable.

\subsection{Other results}
\begin{figure}[t]
	\centering
	\begin{tabular}{@{\hspace{-0.05mm}}c@{\hspace{-0.1mm}}|@{\hspace{-0.05mm}}c@{\hspace{-0.1mm}}|@{\hspace{-0.05mm}}c@{\hspace{-0.1mm}}}
        \includegraphics[width=0.1575\textwidth]{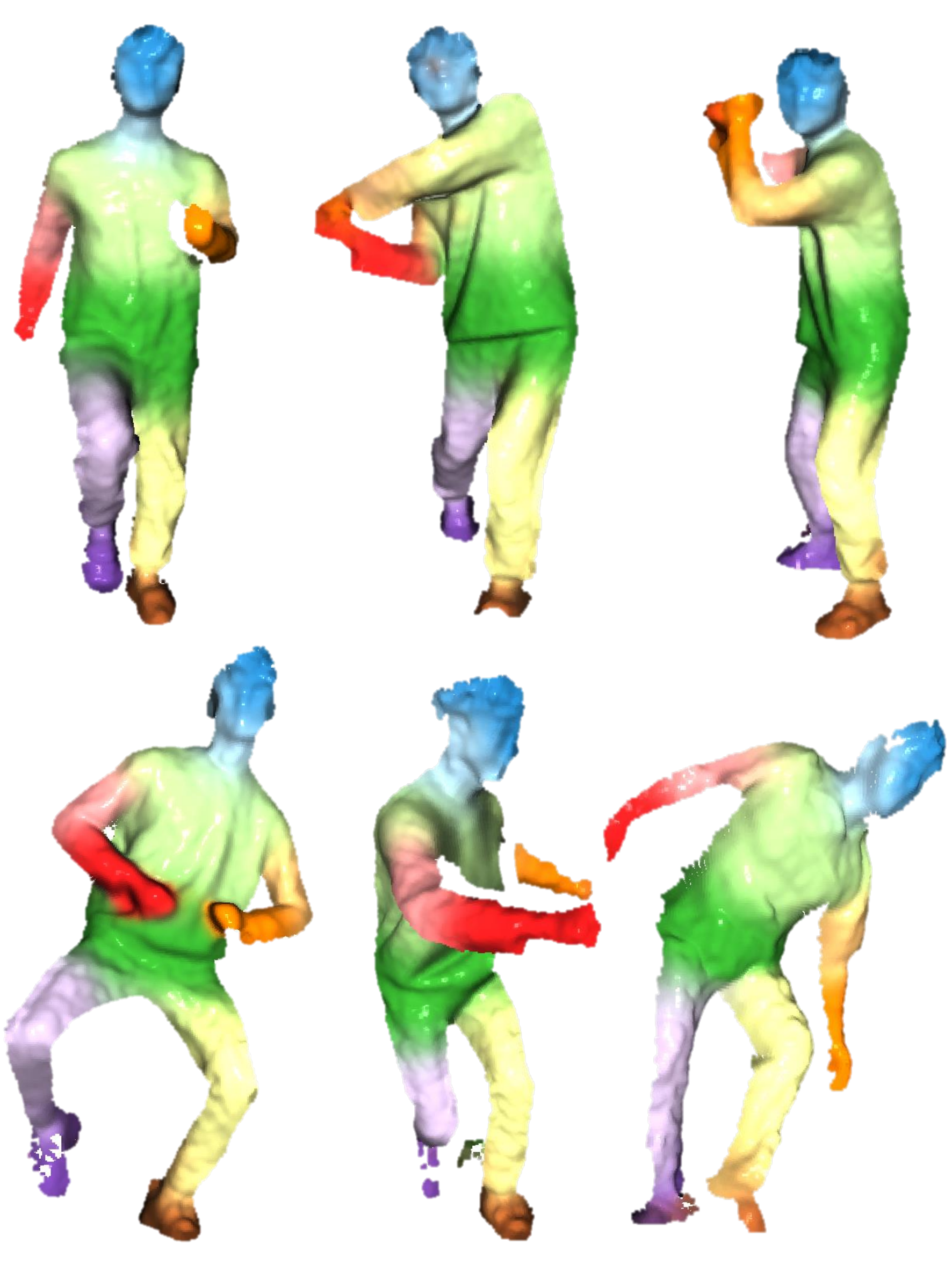}&
        \includegraphics[width=0.158\textwidth]{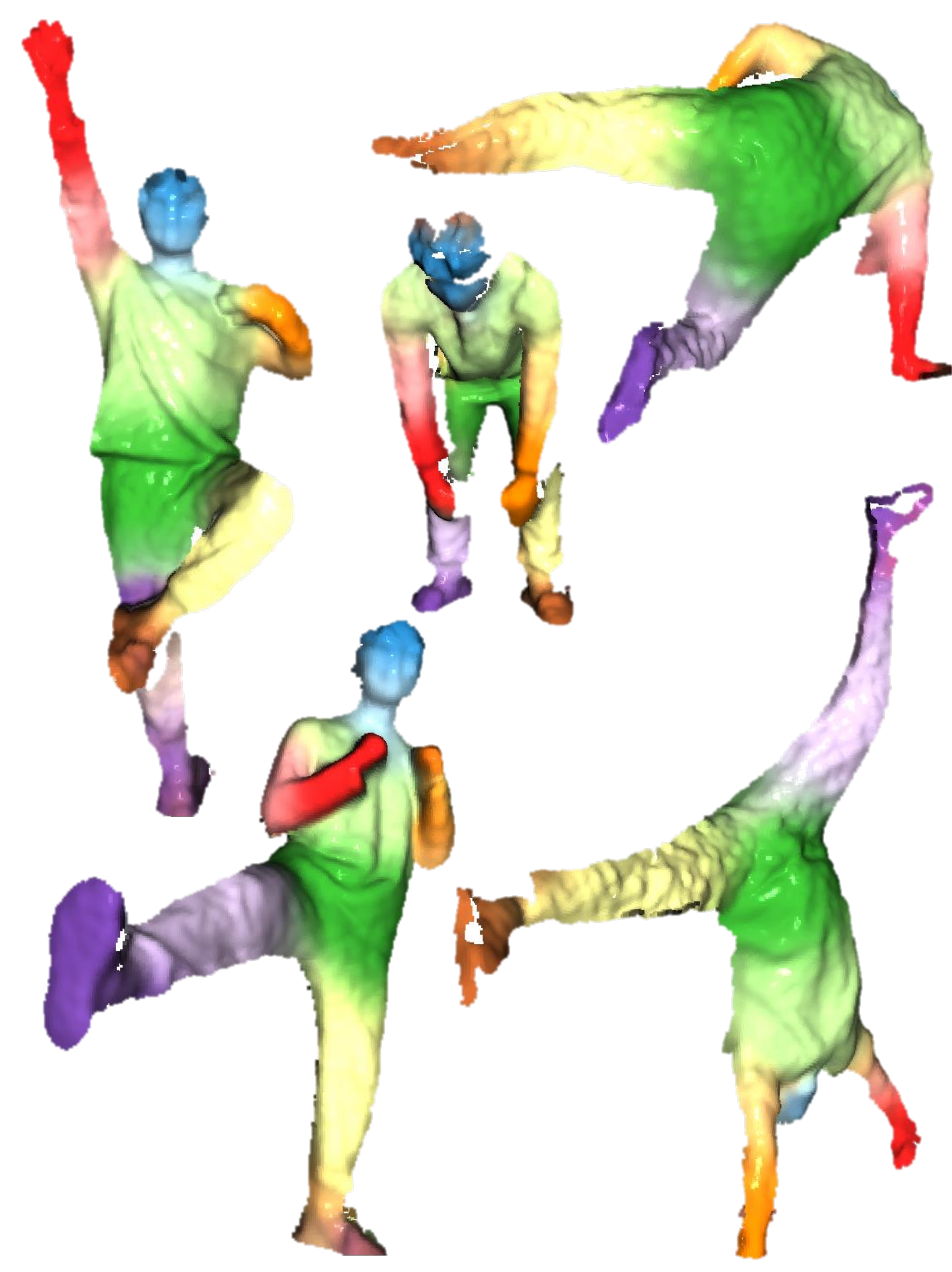}&
        \includegraphics[width=0.1575\textwidth]{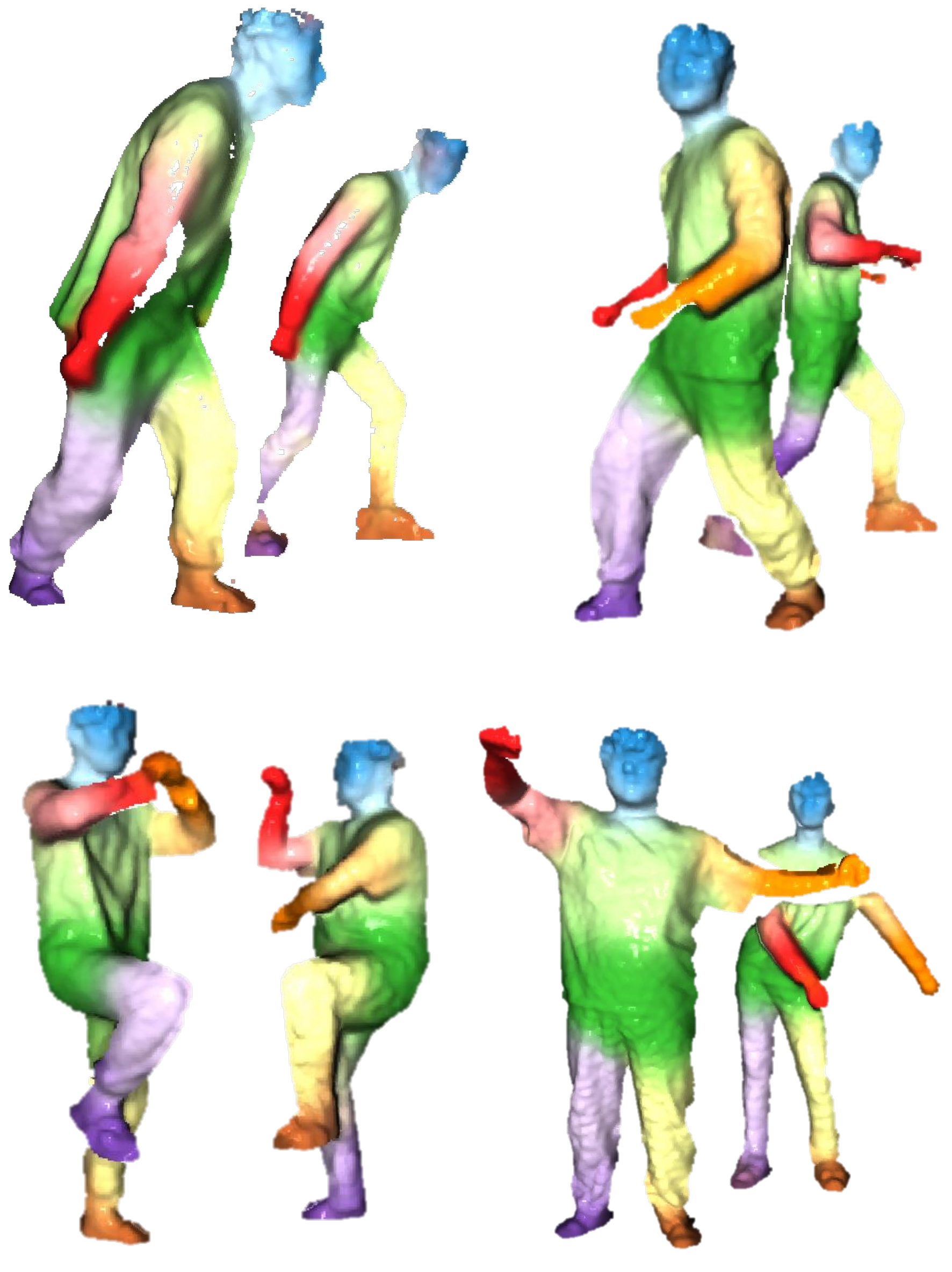}\\
	    \hline
    \end{tabular}
	\begin{tabular}{@{\hspace{-0.05mm}}c@{\hspace{-0.1mm}}@{\hspace{-0.05mm}}c@{\hspace{-0.1mm}}@{\hspace{-0.05mm}}c@{\hspace{-0.01cm}}}
        \includegraphics[width=0.158\textwidth]{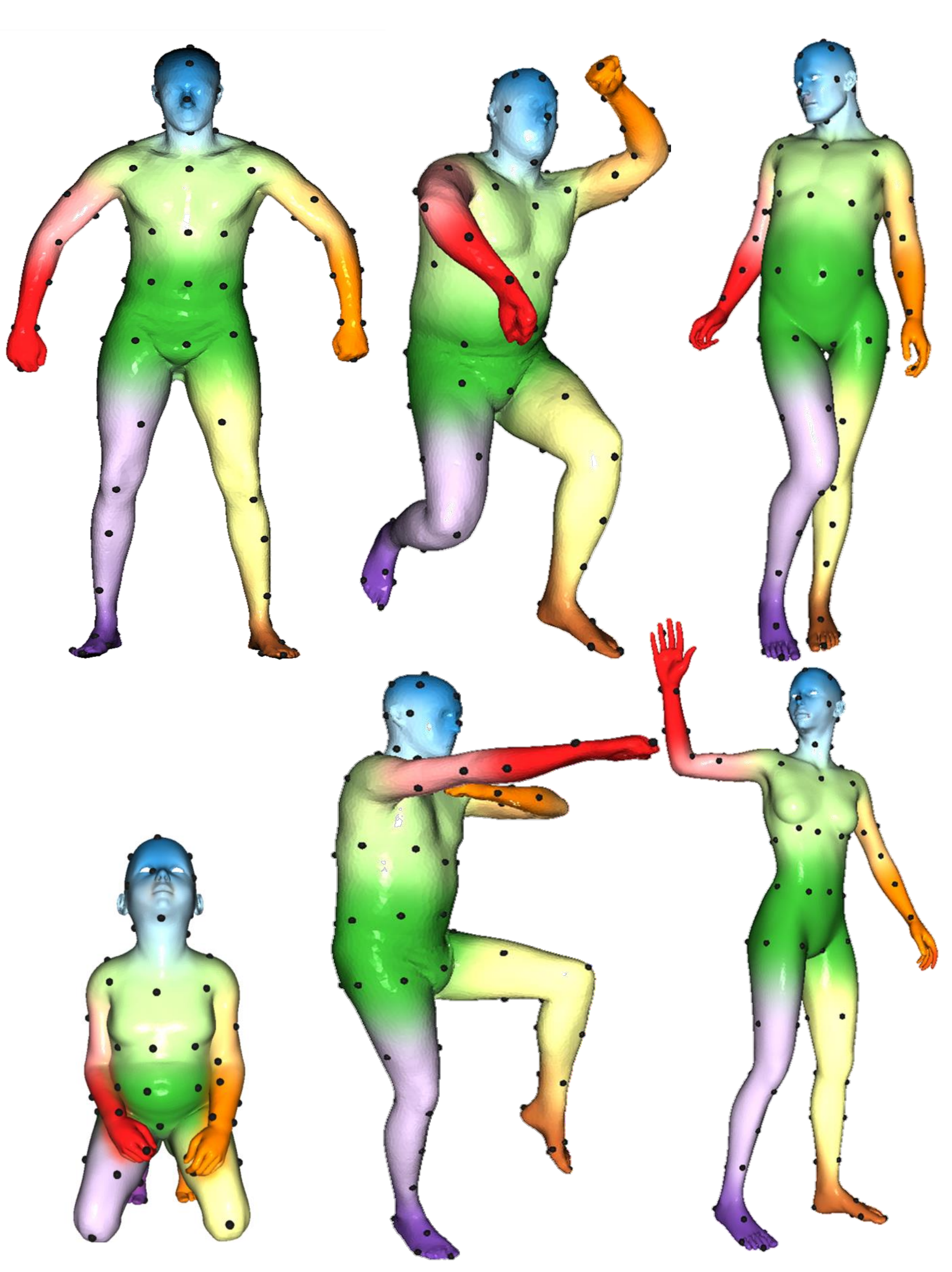}&
        \includegraphics[width=0.158\textwidth]{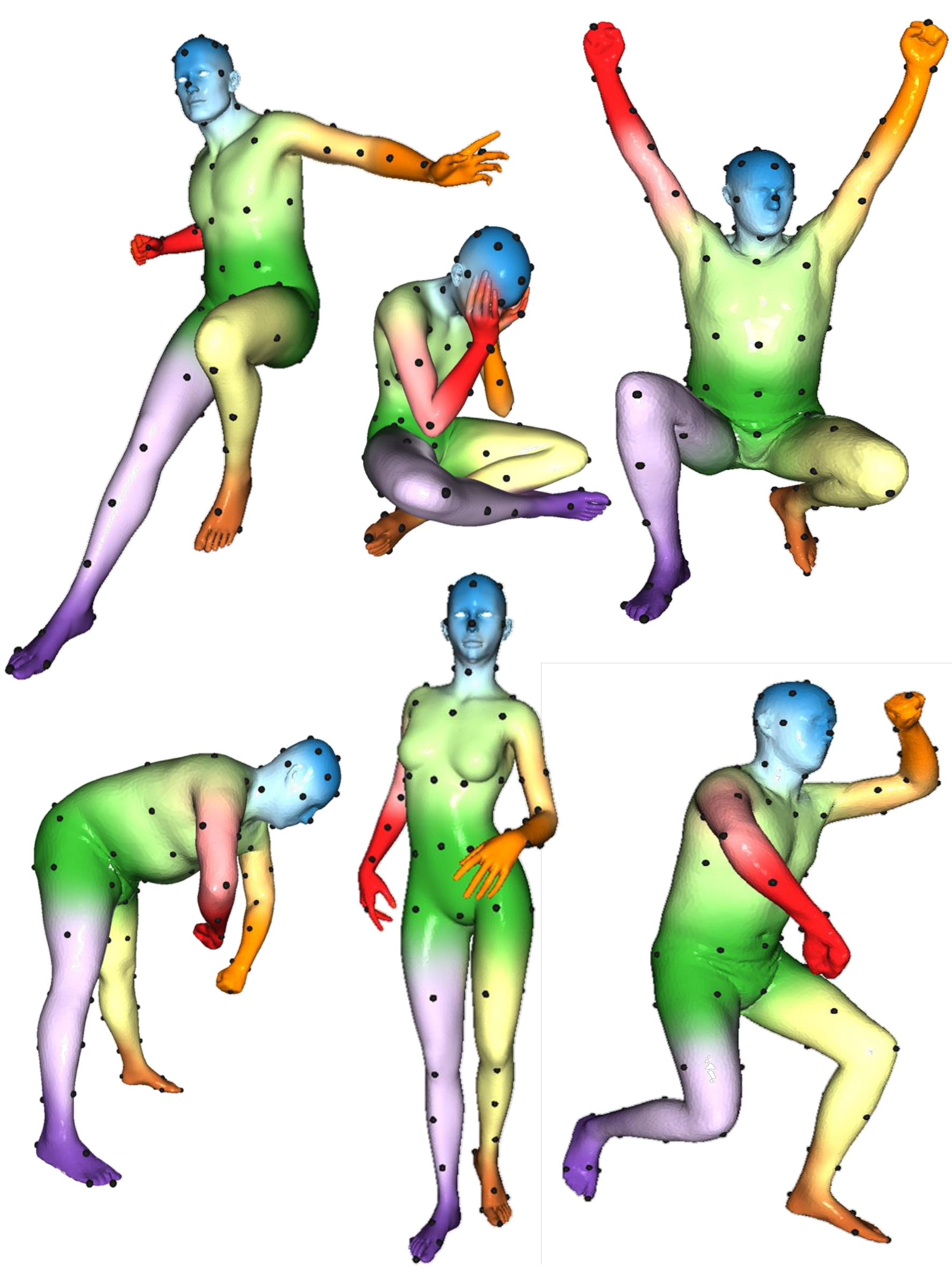}&
        \includegraphics[width=0.158\textwidth]{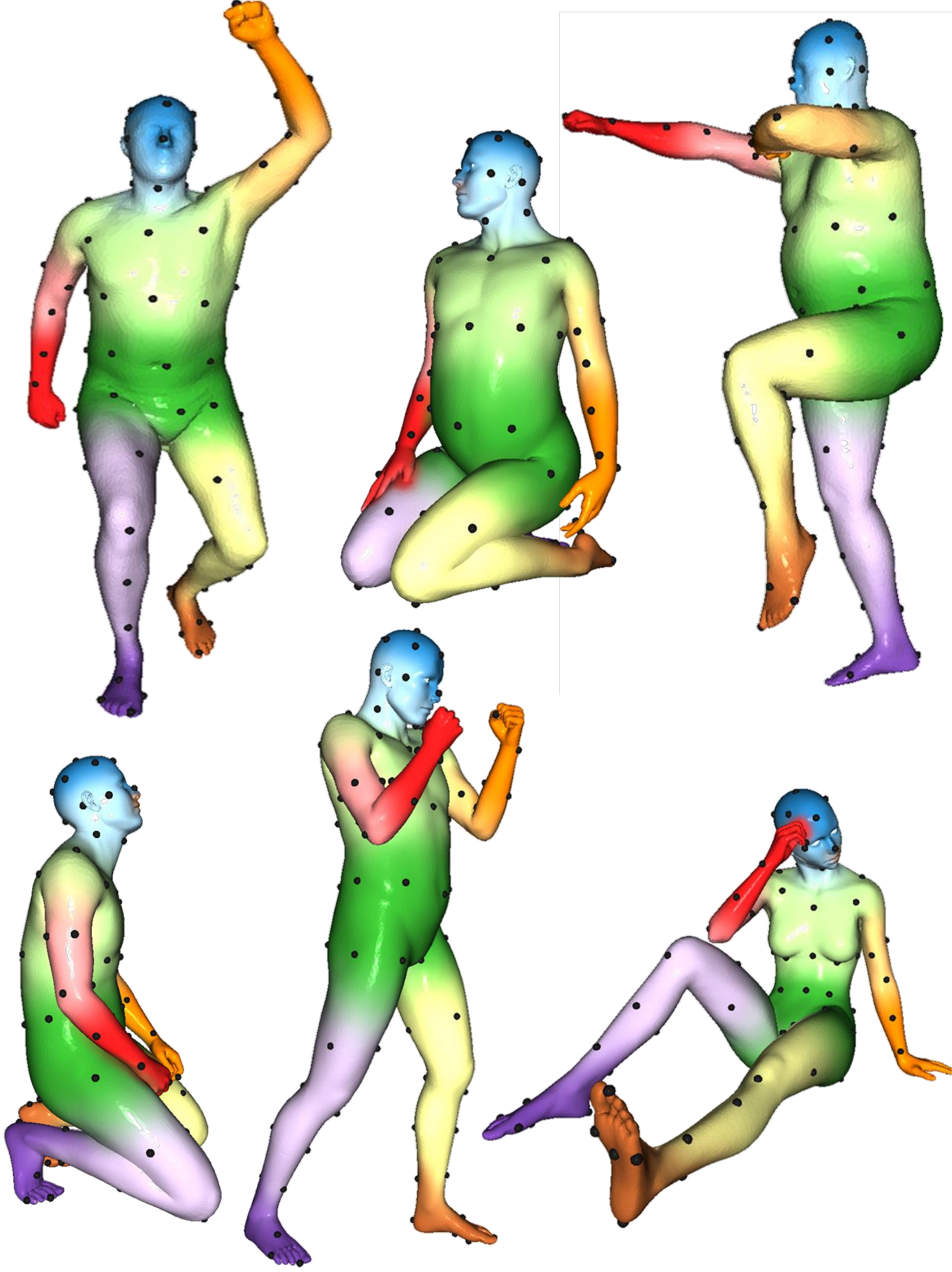}\\
    \end{tabular}
	\caption{Estimated virtual markers in various unseen datasets. (upper part) From left to right, the subfigures show ordinary to challenging cases. The results show that our method is robust to occlusions and diverse poses, and multiple instances. (lower part) Estimated virtual markers on SHREC14~\cite{Pickup2014} dataset. The black dots on the surface indicate estimated sparse virtual markers. To decide the sparse virtual markers, we select the vertex with the highest confidence for each label. The results are consistent with various shapes and motions. Note that we used the training dataset described in Sec.~\ref{sec:dataset}, and these examples are not exposed for the training.
	}
	\vspace*{-7pt}
	\label{fig:exp[qualitative_results]}
\end{figure}

\noindent\textbf{Unseen data.}
We show visual results of our method on two different types of \emph{unseen datasets} - our real depth sequences captured with an Azure Kinect DK~\cite{AzureKinect} and SHREC14~\cite{Pickup2014} dataset. For this experiment, we use the training dataset described in~Sec.~\ref{sec:dataset}.
The upper part of \Fig{exp[qualitative_results]} shows the results on unseen depth data. The depth sequences capture single or multiple people with various motions. Although real depth images are noisy, the results demonstrate that our method is reliable and robust.
The lower part of \Fig{exp[qualitative_results]} showcases color-coded models with our inferenced virtual markers. The results show that our method works well on unseen, various dynamic models.

\begin{figure}[t]
	\centering
	\includegraphics[width=0.48\textwidth]{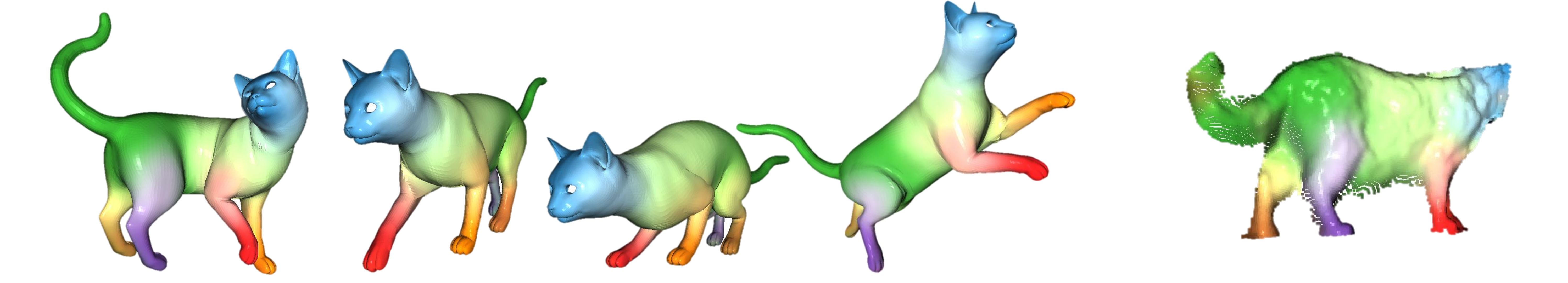} 
	\caption{Example of a cat model. We train the network using 3D cat models obtained by SMAL~\cite{zuffi20173d} with proper random pose data. We then test the network on cat models in TOSCA~\cite{bronstein2008numerical} (the left four results) and a depth map of a real cat (on the right). The reliable results indicate the wide usage of our deep virtual marker.}
	\vspace*{-6pt}
	\label{fig:exp[TOSCA]}
\end{figure}

\vspace{2mm}\noindent\textbf{Cats.}
\Fig{exp[TOSCA]} shows an example for a cat. It shows reasonable predictions on models in TOSCA~\cite{bronstein2008numerical} and a real depth image.


\begin{figure}[t]
	\centering
	\begin{tabular}{c@{\hspace{0.1cm}}c}
		\includegraphics[width=0.18\textwidth]{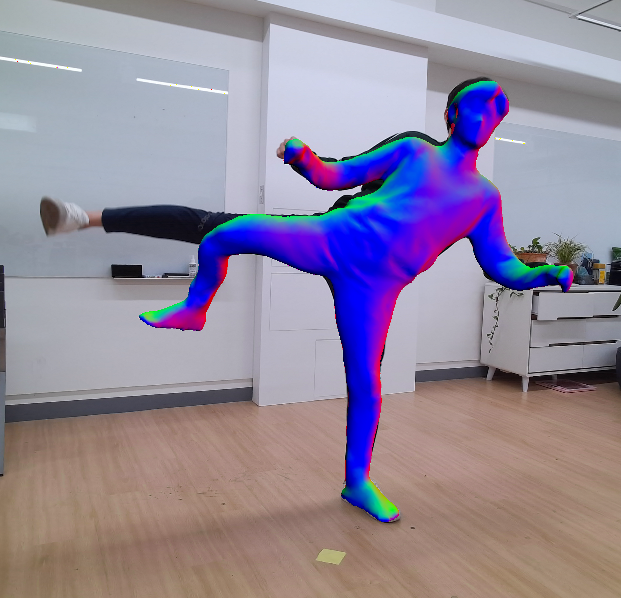} &
		\includegraphics[width=0.18\textwidth]{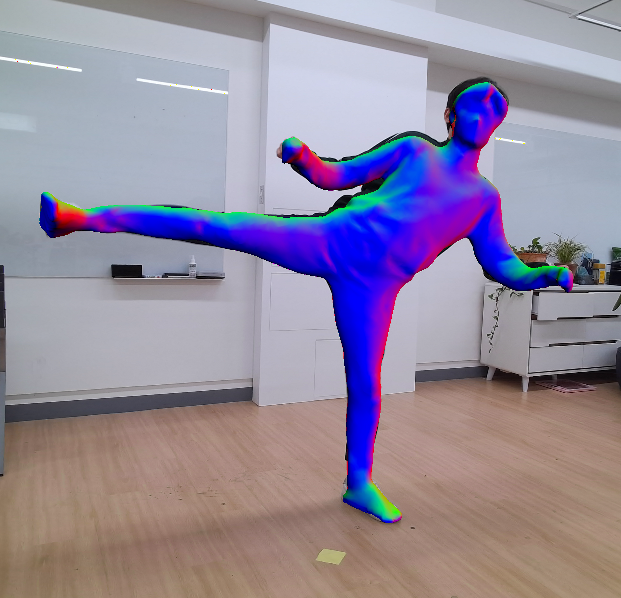}\\
		\small{(a)} & \small{(b)}\\
	\end{tabular}
	\caption{
		Non-rigid registration.
		(a) A large misalignment around a leg due to fast motion cannot be recovered by a local alignment based approach~\cite{li09robust}. (b) Correspondences from our method can successfully remedy such a large misalignment.
	}
	\label{fig:app[nrr]}
\end{figure}

\begin{figure}[t]
	\centering
	\begin{tabular}{c}
		\includegraphics[width=0.48\textwidth]{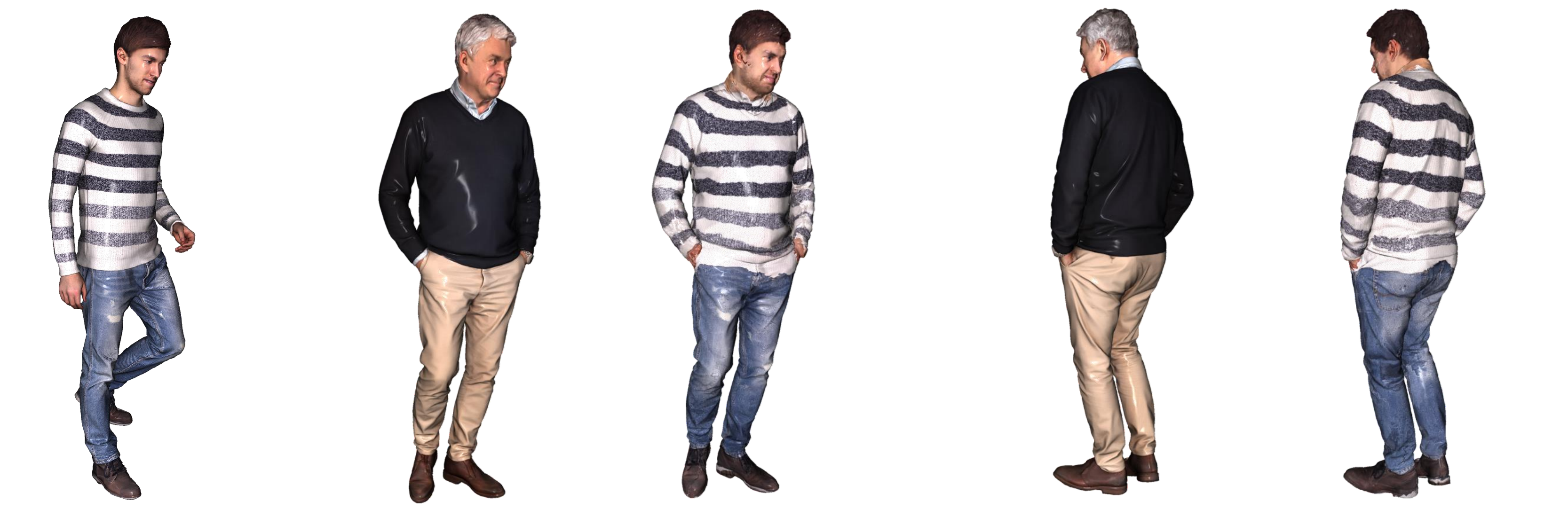}\\
		\hline
		\includegraphics[width=0.48\textwidth]{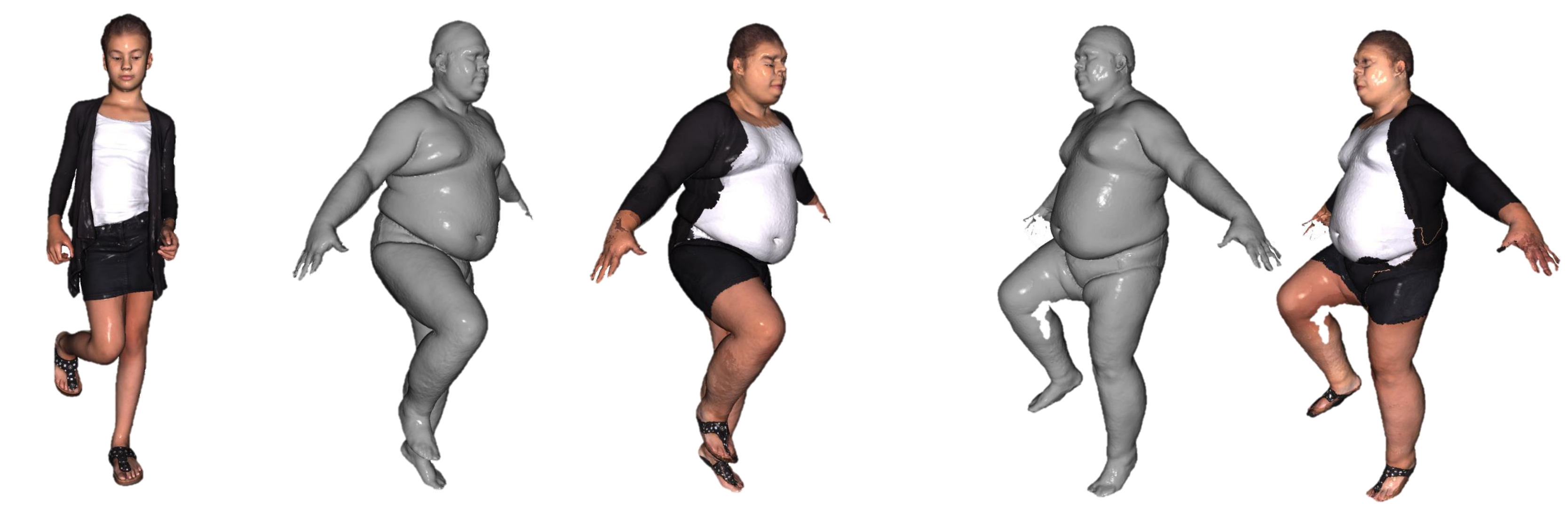}\\
		\hline
		\includegraphics[width=0.48\textwidth]{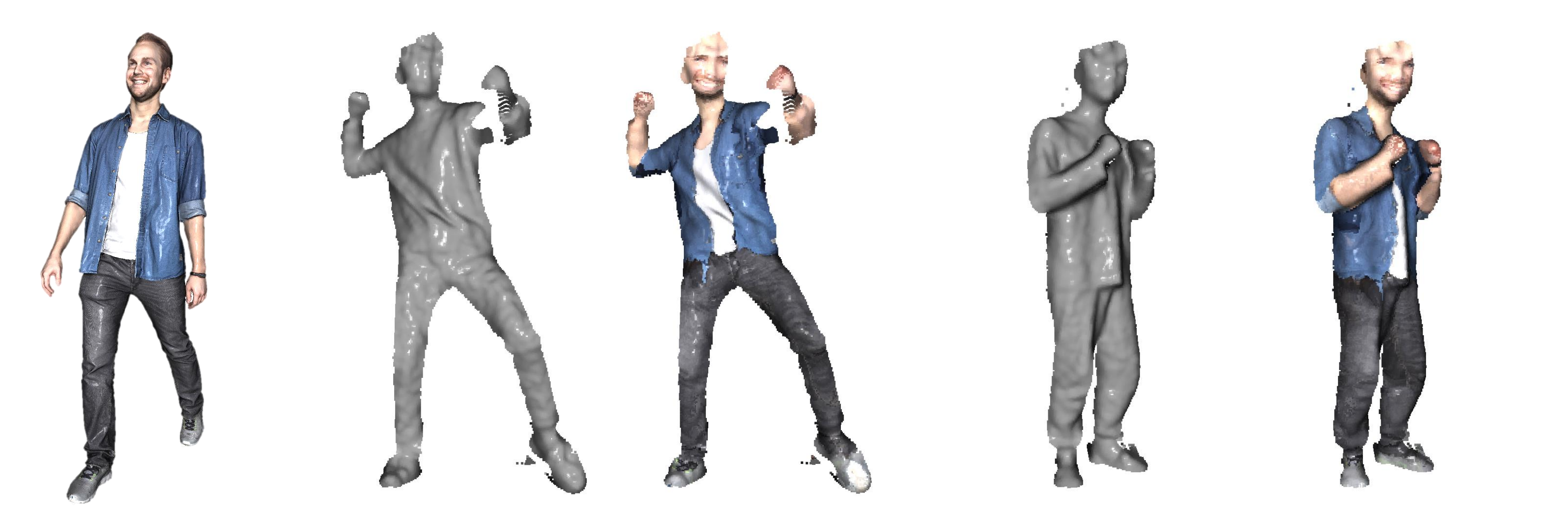}\\
	\end{tabular}
	\caption{
    Texture transfer from a textured mesh to another mesh (first and second row) or a depth map (third row). In the mesh-to-mesh case, we transfer texture coordinates and apply texture mapping to the mesh. In the mesh-to-depth map case, we directly transfer the texture colors. The results show that our deep virtual markers produce sufficiently adequate correspondences even with clothed humans, significant shape differences, or depth noise.
	}
	\label{fig:app[tex]}
\end{figure}

\subsection{Applications}
\noindent\textbf{Non-rigid registration.}
Our approach predicts dense markers, so the correspondence can help to solve non-rigid registration.
One example is shown in \Fig{app[nrr]}. For non-rigid surface registration, we compare ours with Li et al.'s approach~\cite{li09robust} that is a popular baseline for performance capture approaches~\cite{li20133d, dou2016fusion4d, motion2fusion, li20133d, yu2017bodyfusion}. Li et al.'s approach~\cite{li09robust} progressively deforms a source mesh model to fit to the target one. If pose difference between two mesh models is large, it tends to get stuck at local minima since the method uses neighbor search for constructing vertex correspondences, as shown in \Fig{app[nrr]}~(a). The misalignment issue can be alleviated by using dense vertex correspondences estimated by our approach, as shown in \Fig{app[nrr]}~(b).

\vspace{2mm}
\noindent\textbf{Texture transfer.}
As another application, we present texture transfer. By using dense correspondences obtained by our deep virtual markers, the texture in a source mesh model can be instantly mapped to the target mesh model. Similarly, texture transfer can be performed between a full mesh and a depth map, as our deep virtaul markers can handle a variety of 3D inputs in a consistent way. \Fig{app[tex]} shows examples.

\begin{figure}[!t]
    \centering
	\begin{tabular}{c | c | c}
        \includegraphics[width=0.14\textwidth]{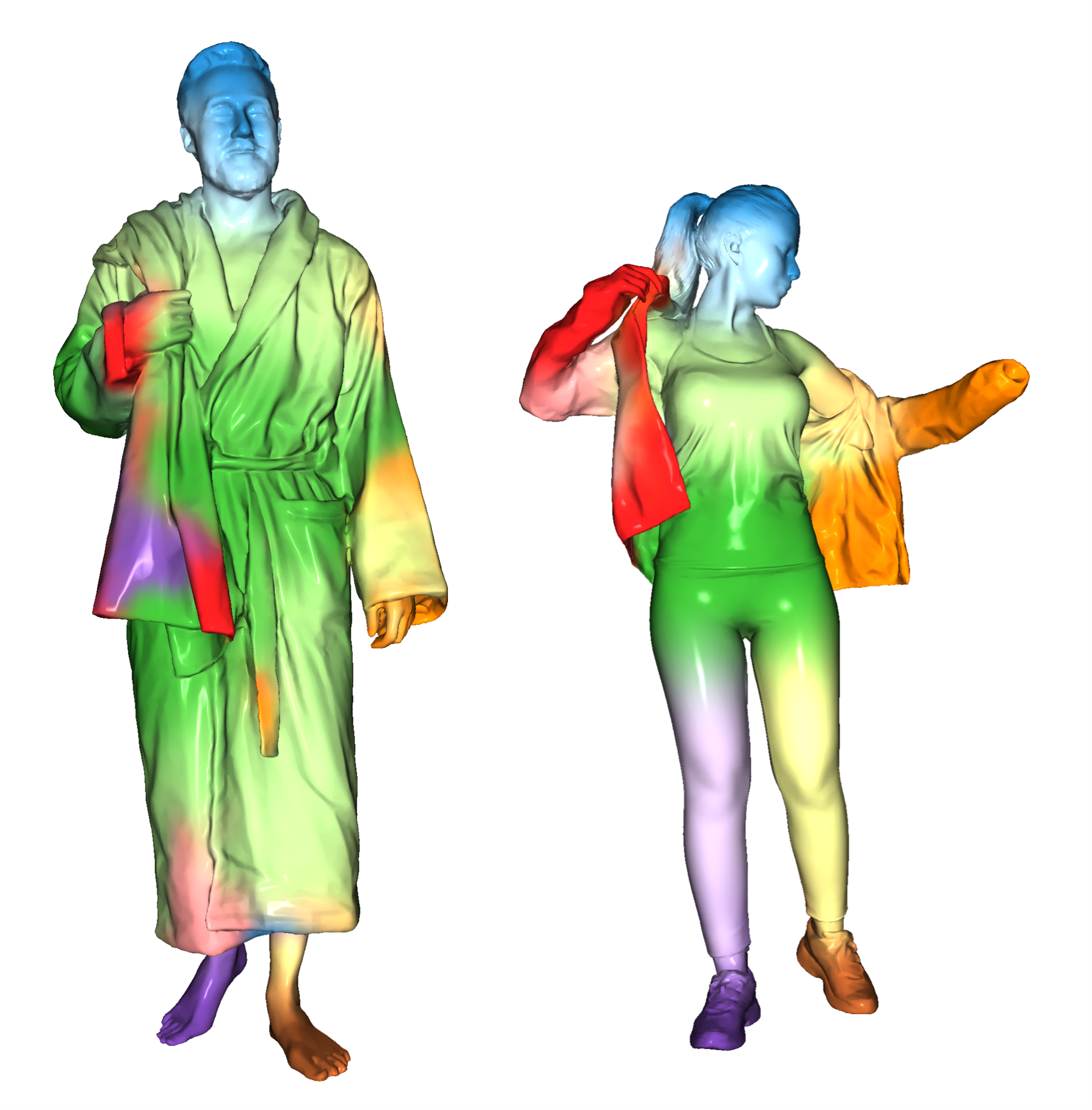}&
        \includegraphics[width=0.14\textwidth]{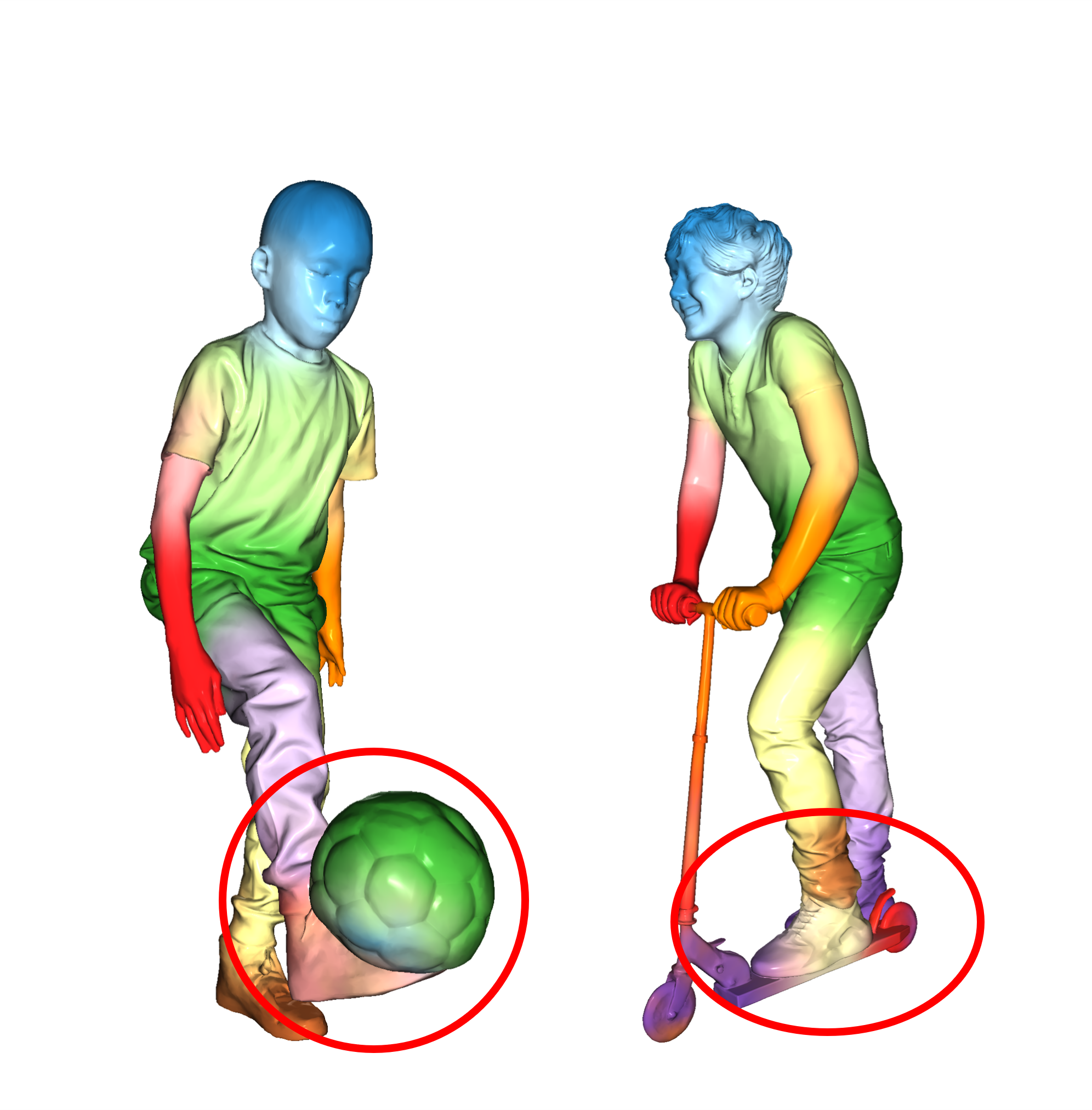}&
        \includegraphics[width=0.14\textwidth]{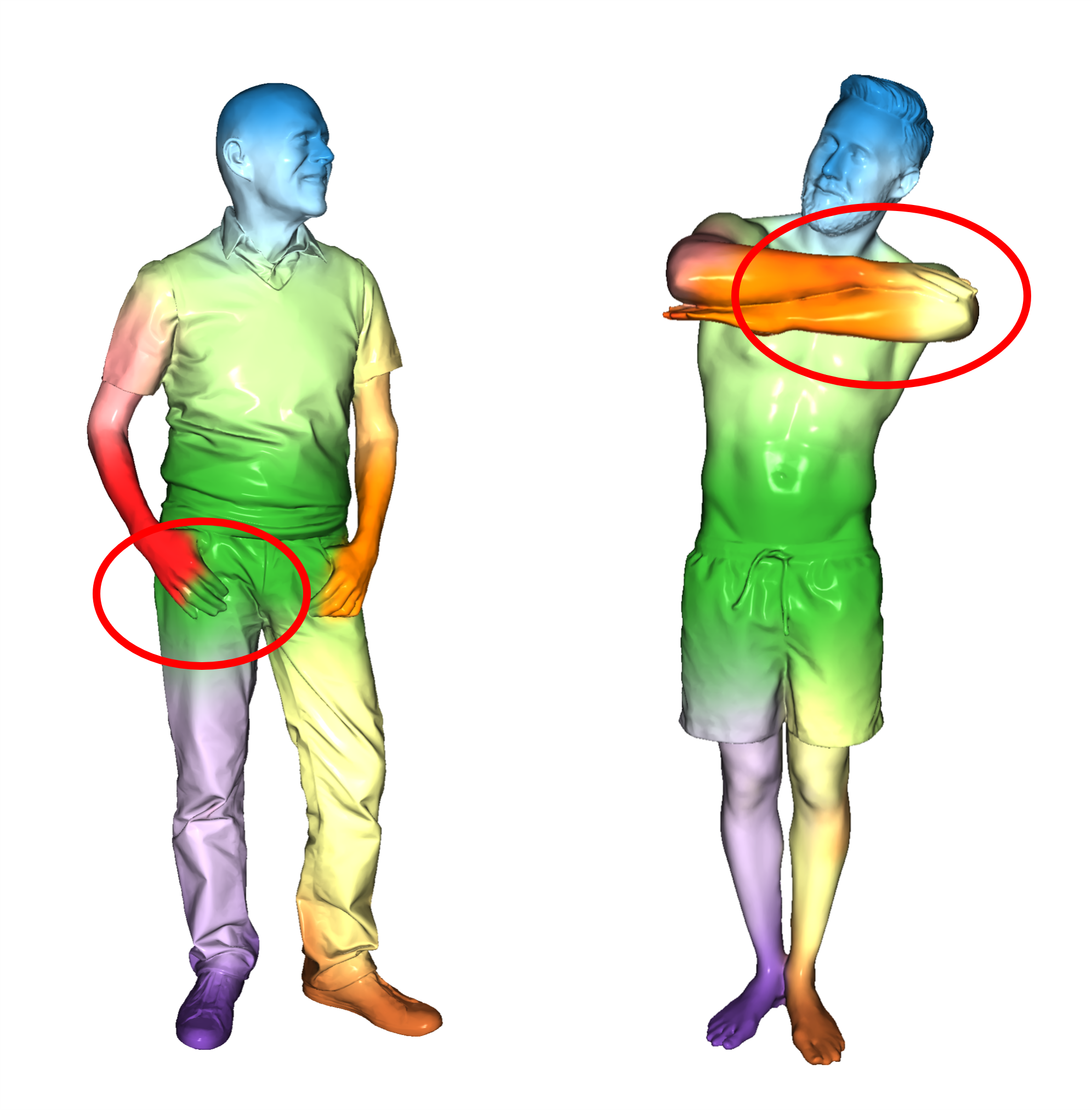}\\
    \end{tabular}
	\caption{Failure cases of our approach. The cases of loose clothes, body-to-object interactions, and intra-body close interactions are shown from left to right.}
	\vspace{-6pt}
	\label{fig:exp[fail_textrans]}
\end{figure}
\subsection{Limitations}
Although our approach is validated with many standard 3D human models from FAUST~\cite{FAUST}, SCAPE~\cite{SCAPE}, and SHREC14~\cite{Pickup2014}, 
our virtual markers may not deal with humans wearing highly loose clothes, such as long skirts and coats. In addition, our method is not very accurate for the cases of body-to-object and intra-body close interactions, as shown in \Fig{exp[fail_textrans]}. We may consider a data-driven solution, but our current pipeline for preparing a dataset requires human annotations, which is not directly applicable to people wearing challenging clothes. Human models interacting with objects are not tested as well. Future work would be to tackle these issues.
\section{Conclusions}
In this paper, we proposed \emph{deep virtual markers}, a methodology that infers dense virtual marker labels for articulated 3D objects. Our approach can handle various types of input, including meshes, point clouds, and depth images. We proposed an effective method to annotate dense virtual markers on 3D meshes to build a training dataset.
Our framework utilizes a fully convolutional neural network as a soft label classifier, and our-oneshot model applies single feed-forward operation to instantly assign dense markers ($0.05 \sim 0.07$ seconds). Compared to state-of-the-art methods, we showed a favorable performance in finding correspondences. Lastly, we demonstrated practical applications such as texture transfer and non-rigid surface registration.
Our future work includes dealing with humans wearing highly loose clothes, such as long skirts and coats.

\paragraph{Acknowledgements}
This work was supported by the IITP grants (SW Star Lab: 2015-0-00174 and Artificial Intelligence Graduate School Program (POSTECH): 2019-0-01906) and the NRF grant (NRF-2020R1C1C1015260) from the Ministry of Science and ICT (MSIT), and the KOCCA grant (R2021040136) from the Ministry of Culture, Sports, and Tourism (MCST).

\balance 


{\small\begin{thebibliography}{10}\itemsep=-1pt

\bibitem{AzureKinect}
Azure kinect dk.
\newblock \url{https://azure.microsoft.com/en-us/services/kinect-dk/}.

\bibitem{mixamo}
Mixamo-3d animation online services, 3d characters, and character rigging.
\newblock \url{https://www.mixamo.com/}.

\bibitem{renderpeople}
Renderpeople-scanned 3d people models provider.
\newblock \url{http://renderpeople.com}.

\bibitem{SCAPE}
Dragomir Anguelov, Praveen Srinivasan, Daphne Koller, Sebastian Thrun, Jim
  Rodgers, and James Davis.
\newblock Scape: Shape completion and animation of people.
\newblock {\em ACM Trans. Graph.}, 24(3):408–416, July 2005.

\bibitem{aubry2011wave}
Mathieu Aubry, Ulrich Schlickewei, and Daniel Cremers.
\newblock The wave kernel signature: A quantum mechanical approach to shape
  analysis.
\newblock In {\em 2011 IEEE international conference on computer vision
  workshops (ICCV workshops)}, pages 1626--1633. IEEE, 2011.

\bibitem{baran2007automatic}
Ilya Baran and Jovan Popovi{\'c}.
\newblock Automatic rigging and animation of 3d characters.
\newblock {\em ACM Transactions on graphics (TOG)}, 26(3):72--es, 2007.

\bibitem{bhatnagar2020loopreg}
Bharat~Lal Bhatnagar, Cristian Sminchisescu, Christian Theobalt, and Gerard
  Pons-Moll.
\newblock Loopreg: Self-supervised learning of implicit surface
  correspondences, pose and shape for 3d human mesh registration.
\newblock {\em Advances in Neural Information Processing Systems}, 33, 2020.

\bibitem{bogo2016keep}
Federica Bogo, Angjoo Kanazawa, Christoph Lassner, Peter Gehler, Javier Romero,
  and Michael~J Black.
\newblock Keep it smpl: Automatic estimation of 3d human pose and shape from a
  single image.
\newblock In {\em European conference on computer vision}, pages 561--578.
  Springer, 2016.

\bibitem{FAUST}
Federica Bogo, Javier Romero, Matthew Loper, and Michael~J. Black.
\newblock {FAUST}: Dataset and evaluation for {3D} mesh registration.
\newblock In {\em Proceedings IEEE Conf. on Computer Vision and Pattern
  Recognition (CVPR)}, Piscataway, NJ, USA, June 2014. IEEE.

\bibitem{boscaini2016learning}
Davide Boscaini, Jonathan Masci, Emanuele Rodoi{\`a}, and Michael Bronstein.
\newblock Learning shape correspondence with anisotropic convolutional neural
  networks.
\newblock In {\em Proceedings of the 30th International Conference on Neural
  Information Processing Systems}, pages 3197--3205, 2016.

\bibitem{bregler2007motion}
Chris Bregler.
\newblock Motion capture technology for entertainment [in the spotlight].
\newblock {\em IEEE Signal Processing Magazine}, 24(6):160--158, 2007.

\bibitem{bronstein2008numerical}
Alexander~M Bronstein, Michael~M Bronstein, and Ron Kimmel.
\newblock {\em Numerical geometry of non-rigid shapes}.
\newblock Springer Science \& Business Media, 2008.

\bibitem{chan2010virtual}
Jacky~CP Chan, Howard Leung, Jeff~KT Tang, and Taku Komura.
\newblock A virtual reality dance training system using motion capture
  technology.
\newblock {\em IEEE transactions on learning technologies}, 4(2):187--195,
  2010.

\bibitem{chen2015robust}
Qifeng Chen and Vladlen Koltun.
\newblock Robust nonrigid registration by convex optimization.
\newblock In {\em Proceedings of the IEEE International Conference on Computer
  Vision}, pages 2039--2047, 2015.

\bibitem{choy20194d}
Christopher Choy, JunYoung Gwak, and Silvio Savarese.
\newblock 4d spatio-temporal convnets: Minkowski convolutional neural networks.
\newblock In {\em Proceedings of the IEEE Conference on Computer Vision and
  Pattern Recognition}, pages 3075--3084, 2019.

\bibitem{choy2019fully}
Christopher Choy, Jaesik Park, and Vladlen Koltun.
\newblock Fully convolutional geometric features.
\newblock In {\em Proceedings of the IEEE/CVF International Conference on
  Computer Vision}, pages 8958--8966, 2019.

\bibitem{deprelle2019learning}
Theo Deprelle, Thibault Groueix, Matthew Fisher, Vladimir~G Kim, Bryan~C
  Russell, and Mathieu Aubry.
\newblock Learning elementary structures for 3d shape generation and matching.
\newblock {\em arXiv preprint arXiv:1908.04725}, 2019.

\bibitem{donati2020deep}
Nicolas Donati, Abhishek Sharma, and Maks Ovsjanikov.
\newblock Deep geometric functional maps: Robust feature learning for shape
  correspondence.
\newblock In {\em Proceedings of the IEEE/CVF Conference on Computer Vision and
  Pattern Recognition}, pages 8592--8601, 2020.

\bibitem{motion2fusion}
Mingsong Dou, Philip Davidson, Sean~Ryan Fanello, Sameh Khamis, Adarsh Kowdle,
  Christoph Rhemann, Vladimir Tankovich, and Shahram Izadi.
\newblock Motion2fusion: Real-time volumetric performance capture.
\newblock 36(6), Nov. 2017.

\bibitem{dou2016fusion4d}
Mingsong Dou, Sameh Khamis, Yury Degtyarev, Philip Davidson, Sean~Ryan Fanello,
  Adarsh Kowdle, Sergio~Orts Escolano, Christoph Rhemann, David Kim, Jonathan
  Taylor, et~al.
\newblock Fusion4d: Real-time performance capture of challenging scenes.
\newblock {\em ACM Transactions on Graphics (TOG)}, 35(4):1--13, 2016.

\bibitem{eisenberger2020smooth}
Marvin Eisenberger, Zorah Lahner, and Daniel Cremers.
\newblock Smooth shells: Multi-scale shape registration with functional maps.
\newblock In {\em Proceedings of the IEEE/CVF Conference on Computer Vision and
  Pattern Recognition}, pages 12265--12274, 2020.

\bibitem{eisenberger2020deep}
Marvin Eisenberger, Aysim Toker, Laura Leal-Taix{\'e}, and Daniel Cremers.
\newblock Deep shells: Unsupervised shape correspondence with optimal
  transport.
\newblock {\em arXiv preprint arXiv:2010.15261}, 2020.

\bibitem{Fey_2018_CVPR}
Matthias Fey, Jan~Eric Lenssen, Frank Weichert, and Heinrich Müller.
\newblock Splinecnn: Fast geometric deep learning with continuous b-spline
  kernels.
\newblock In {\em Proceedings of the IEEE Conference on Computer Vision and
  Pattern Recognition (CVPR)}, June 2018.

\bibitem{frome2004recognizing}
Andrea Frome, Daniel Huber, Ravi Kolluri, Thomas B{\"u}low, and Jitendra Malik.
\newblock Recognizing objects in range data using regional point descriptors.
\newblock In {\em European conference on computer vision}, pages 224--237.
  Springer, 2004.

\bibitem{gao2020learning}
Zhongpai Gao, Guangtao Zhai, Juyong Zhang, Junchi Yan, Yiyan Yang, and Xiaokang
  Yang.
\newblock Learning local neighboring structure for robust 3d shape
  representation.
\newblock {\em arXiv e-prints}, pages arXiv--2004, 2020.

\bibitem{ginzburg2020cyclic}
Dvir Ginzburg and Dan Raviv.
\newblock Cyclic functional mapping: Self-supervised correspondence between
  non-isometric deformable shapes.
\newblock In {\em European Conference on Computer Vision}, pages 36--52.
  Springer, 2020.

\bibitem{Gong_2019_ICCV}
Shunwang Gong, Lei Chen, Michael Bronstein, and Stefanos Zafeiriou.
\newblock Spiralnet++: A fast and highly efficient mesh convolution operator.
\newblock In {\em Proceedings of the IEEE/CVF International Conference on
  Computer Vision (ICCV) Workshops}, Oct 2019.

\bibitem{groueix20183d}
Thibault Groueix, Matthew Fisher, Vladimir~G Kim, Bryan~C Russell, and Mathieu
  Aubry.
\newblock 3d-coded: 3d correspondences by deep deformation.
\newblock In {\em Proceedings of the European Conference on Computer Vision
  (ECCV)}, pages 230--246, 2018.

\bibitem{habermann2019livecap}
Marc Habermann, Weipeng Xu, Michael Zollhoefer, Gerard Pons-Moll, and Christian
  Theobalt.
\newblock Livecap: Real-time human performance capture from monocular video.
\newblock {\em ACM Transactions On Graphics (TOG)}, 38(2):1--17, 2019.

\bibitem{habermann2020deepcap}
Marc Habermann, Weipeng Xu, Michael Zollhofer, Gerard Pons-Moll, and Christian
  Theobalt.
\newblock Deepcap: Monocular human performance capture using weak supervision.
\newblock In {\em Proceedings of the IEEE/CVF Conference on Computer Vision and
  Pattern Recognition}, pages 5052--5063, 2020.

\bibitem{halimi2019unsupervised}
Oshri Halimi, Or Litany, Emanuele Rodola, Alex~M Bronstein, and Ron Kimmel.
\newblock Unsupervised learning of dense shape correspondence.
\newblock In {\em Proceedings of the IEEE/CVF Conference on Computer Vision and
  Pattern Recognition}, pages 4370--4379, 2019.

\bibitem{hernandez2012graph}
Antonio Hern{\'a}ndez-Vela, Nadezhda Zlateva, Alexander Marinov, Miguel Reyes,
  Petia Radeva, Dimo Dimov, and Sergio Escalera.
\newblock Graph cuts optimization for multi-limb human segmentation in depth
  maps.
\newblock In {\em 2012 IEEE Conference on Computer Vision and Pattern
  Recognition}, pages 726--732. IEEE, 2012.

\bibitem{johnson1999using}
Andrew~E. Johnson and Martial Hebert.
\newblock Using spin images for efficient object recognition in cluttered 3d
  scenes.
\newblock {\em IEEE Transactions on pattern analysis and machine intelligence},
  21(5):433--449, 1999.

\bibitem{Joo_2018_CVPR}
Hanbyul Joo, Tomas Simon, and Yaser Sheikh.
\newblock Total capture: A 3d deformation model for tracking faces, hands, and
  bodies.
\newblock In {\em Proceedings of the IEEE Conference on Computer Vision and
  Pattern Recognition (CVPR)}, June 2018.

\bibitem{kavan2007skinning}
Ladislav Kavan, Steven Collins, Ji{\v{r}}{\'\i} {\v{Z}}{\'a}ra, and Carol
  O'Sullivan.
\newblock Skinning with dual quaternions.
\newblock In {\em Proceedings of the 2007 symposium on Interactive 3D graphics
  and games}, pages 39--46, 2007.

\bibitem{Kim2011BIM}
Vladimir Kim, Yaron Lipman, and Thomas Funkhouser.
\newblock Blended intrinsic maps.
\newblock {\em ACM Transactions on Graphics (Proc. SIGGRAPH)}, 30(4), July
  2011.

\bibitem{lassner2017unite}
Christoph Lassner, Javier Romero, Martin Kiefel, Federica Bogo, Michael~J
  Black, and Peter~V Gehler.
\newblock Unite the people: Closing the loop between 3d and 2d human
  representations.
\newblock In {\em Proceedings of the IEEE conference on computer vision and
  pattern recognition}, pages 6050--6059, 2017.

\bibitem{li2019lbs}
Chun-Liang Li, Tomas Simon, Jason Saragih, Barnab{\'a}s P{\'o}czos, and Yaser
  Sheikh.
\newblock Lbs autoencoder: Self-supervised fitting of articulated meshes to
  point clouds.
\newblock In {\em Proceedings of the IEEE/CVF Conference on Computer Vision and
  Pattern Recognition}, pages 11967--11976, 2019.

\bibitem{li09robust}
Hao Li, Bart Adams, Leonidas~J. Guibas, and Mark Pauly.
\newblock Robust single-view geometry and motion reconstruction.
\newblock {\em ACM Transactions on Graphics (Proceedings SIGGRAPH Asia 2009)},
  28(5), December 2009.

\bibitem{li20133d}
Hao Li, Etienne Vouga, Anton Gudym, Linjie Luo, Jonathan~T Barron, and Gleb
  Gusev.
\newblock 3d self-portraits.
\newblock {\em ACM Transactions on Graphics (TOG)}, 32(6):1--9, 2013.

\bibitem{lim2018simple}
Isaak Lim, Alexander Dielen, Marcel Campen, and Leif Kobbelt.
\newblock A simple approach to intrinsic correspondence learning on
  unstructured 3d meshes.
\newblock In {\em Proceedings of the European Conference on Computer Vision
  (ECCV) Workshops}, pages 0--0, 2018.

\bibitem{litany2017deep}
Or Litany, Tal Remez, Emanuele Rodola, Alex Bronstein, and Michael Bronstein.
\newblock Deep functional maps: Structured prediction for dense shape
  correspondence.
\newblock In {\em Proceedings of the IEEE international conference on computer
  vision}, pages 5659--5667, 2017.

\bibitem{marin2020farm}
Riccardo Marin, Simone Melzi, Emanuele Rodola, and Umberto Castellani.
\newblock Farm: Functional automatic registration method for 3d human bodies.
\newblock In {\em Computer Graphics Forum}, volume~39, pages 160--173. Wiley
  Online Library, 2020.

\bibitem{marin2020correspondence}
Riccardo Marin, Marie-Julie Rakotosaona, Simone Melzi, and Maks Ovsjanikov.
\newblock Correspondence learning via linearly-invariant embedding.
\newblock {\em Advances in Neural Information Processing Systems}, 33, 2020.

\bibitem{masci2015geodesic}
Jonathan Masci, Davide Boscaini, Michael Bronstein, and Pierre Vandergheynst.
\newblock Geodesic convolutional neural networks on riemannian manifolds.
\newblock In {\em Proceedings of the IEEE international conference on computer
  vision workshops}, pages 37--45, 2015.

\bibitem{melzi2019zoomout}
Simone Melzi, Jing Ren, Emanuele Rodola, Abhishek Sharma, Peter Wonka, and Maks
  Ovsjanikov.
\newblock Zoomout: Spectral upsampling for efficient shape correspondence.
\newblock {\em arXiv preprint arXiv:1904.07865}, 2019.

\bibitem{menache2000understanding}
Alberto Menache.
\newblock {\em Understanding motion capture for computer animation and video
  games}.
\newblock Morgan kaufmann, 2000.

\bibitem{MOESLUND2001231}
Thomas~B. Moeslund and Erik Granum.
\newblock A survey of computer vision-based human motion capture.
\newblock {\em Computer Vision and Image Understanding}, 81(3):231--268, 2001.

\bibitem{Monti_2017_CVPR}
Federico Monti, Davide Boscaini, Jonathan Masci, Emanuele Rodola, Jan Svoboda,
  and Michael~M. Bronstein.
\newblock Geometric deep learning on graphs and manifolds using mixture model
  cnns.
\newblock In {\em Proceedings of the IEEE Conference on Computer Vision and
  Pattern Recognition (CVPR)}, July 2017.

\bibitem{mu2020learning}
Jiteng Mu, Weichao Qiu, Gregory~D Hager, and Alan~L Yuille.
\newblock Learning from synthetic animals.
\newblock In {\em Proceedings of the IEEE/CVF Conference on Computer Vision and
  Pattern Recognition}, pages 12386--12395, 2020.

\bibitem{newell2016stacked}
Alejandro Newell, Kaiyu Yang, and Jia Deng.
\newblock Stacked hourglass networks for human pose estimation.
\newblock In {\em European conference on computer vision}, pages 483--499.
  Springer, 2016.

\bibitem{NISHI2017402}
K. Nishi and J. Miura.
\newblock Generation of human depth images with body part labels for complex
  human pose recognition.
\newblock {\em Pattern Recognition}, 71:402--413, 2017.

\bibitem{oliveira2016deep}
Gabriel~L Oliveira, Abhinav Valada, Claas Bollen, Wolfram Burgard, and Thomas
  Brox.
\newblock Deep learning for human part discovery in images.
\newblock In {\em 2016 IEEE International conference on robotics and automation
  (ICRA)}, pages 1634--1641. IEEE, 2016.

\bibitem{ovsjanikov2012functional}
Maks Ovsjanikov, Mirela Ben-Chen, Justin Solomon, Adrian Butscher, and Leonidas
  Guibas.
\newblock Functional maps: a flexible representation of maps between shapes.
\newblock {\em ACM Transactions on Graphics (TOG)}, 31(4):1--11, 2012.

\bibitem{sigCourseFuntionalMaps}
Maks Ovsjanikov, Etienne Corman, Michael Bronstein, Emanuele Rodol\`{a}, Mirela
  Ben-Chen, Leonidas Guibas, Frederic Chazal, and Alex Bronstein.
\newblock Computing and processing correspondences with functional maps.
\newblock In {\em ACM SIGGRAPH 2017 Courses}, SIGGRAPH '17, New York, NY, USA,
  2017. Association for Computing Machinery.

\bibitem{Pickup2014}
D. Pickup, X. Sun, P.~L. Rosin, R.~R. Martin, Z. Cheng, Z. Lian, M. Aono, A.
  Ben~Hamza, A. Bronstein, M. Bronstein, S. Bu, U. Castellani, S. Cheng, V.
  Garro, A. Giachetti, A. Godil, J. Han, H. Johan, L. Lai, B. Li, C. Li, H. Li,
  R. Litman, X. Liu, Z. Liu, Y. Lu, A. Tatsuma, and J. Ye.
\newblock S{H}{R}{E}{C}'14 track: Shape retrieval of non-rigid 3d human models.
\newblock In {\em Proceedings of the 7th Eurographics workshop on 3D Object
  Retrieval}, EG 3DOR'14. Eurographics Association, 2014.

\bibitem{prokudin2019efficient}
Sergey Prokudin, Christoph Lassner, and Javier Romero.
\newblock Efficient learning on point clouds with basis point sets.
\newblock In {\em Proceedings of the IEEE/CVF International Conference on
  Computer Vision}, pages 4332--4341, 2019.

\bibitem{ren2018continuous}
Jing Ren, Adrien Poulenard, Peter Wonka, and Maks Ovsjanikov.
\newblock Continuous and orientation-preserving correspondences via functional
  maps.
\newblock {\em ACM Transactions on Graphics (ToG)}, 37(6):1--16, 2018.

\bibitem{roufosse2019unsupervised}
Jean-Michel Roufosse, Abhishek Sharma, and Maks Ovsjanikov.
\newblock Unsupervised deep learning for structured shape matching.
\newblock In {\em Proceedings of the IEEE/CVF International Conference on
  Computer Vision}, pages 1617--1627, 2019.

\bibitem{rusu2009fast}
Radu~Bogdan Rusu, Nico Blodow, and Michael Beetz.
\newblock Fast point feature histograms (fpfh) for 3d registration.
\newblock In {\em 2009 IEEE international conference on robotics and
  automation}, pages 3212--3217. IEEE, 2009.

\bibitem{salti2014shot}
Samuele Salti, Federico Tombari, and Luigi Di~Stefano.
\newblock Shot: Unique signatures of histograms for surface and texture
  description.
\newblock {\em Computer Vision and Image Understanding}, 125:251--264, 2014.

\bibitem{PhysCap2020}
Soshi Shimada, Vladislav Golyanik, Weipeng Xu, and Christian Theobalt.
\newblock Physcap: Physically plausible monocular 3d motion capture in real
  time.
\newblock 39(6), 2020.

\bibitem{shotton2011real}
Jamie Shotton, Andrew Fitzgibbon, Mat Cook, Toby Sharp, Mark Finocchio, Richard
  Moore, Alex Kipman, and Andrew Blake.
\newblock Real-time human pose recognition in parts from single depth images.
\newblock In {\em CVPR 2011}, pages 1297--1304. Ieee, 2011.

\bibitem{sun2009concise}
Jian Sun, Maks Ovsjanikov, and Leonidas Guibas.
\newblock A concise and provably informative multi-scale signature based on
  heat diffusion.
\newblock In {\em Computer graphics forum}, volume~28, pages 1383--1392. Wiley
  Online Library, 2009.

\bibitem{Thomas_2019_ICCV}
Hugues Thomas, Charles~R. Qi, Jean-Emmanuel Deschaud, Beatriz Marcotegui,
  Francois Goulette, and Leonidas~J. Guibas.
\newblock Kpconv: Flexible and deformable convolution for point clouds.
\newblock In {\em Proceedings of the IEEE/CVF International Conference on
  Computer Vision (ICCV)}, October 2019.

\bibitem{varol2017learning}
Gul Varol, Javier Romero, Xavier Martin, Naureen Mahmood, Michael~J Black, Ivan
  Laptev, and Cordelia Schmid.
\newblock Learning from synthetic humans.
\newblock In {\em Proceedings of the IEEE Conference on Computer Vision and
  Pattern Recognition}, pages 109--117, 2017.

\bibitem{Verma_2018_CVPR}
Nitika Verma, Edmond Boyer, and Jakob Verbeek.
\newblock Feastnet: Feature-steered graph convolutions for 3d shape analysis.
\newblock In {\em Proceedings of the IEEE Conference on Computer Vision and
  Pattern Recognition (CVPR)}, June 2018.

\bibitem{wang2015semantic}
Jianyu Wang and Alan~L Yuille.
\newblock Semantic part segmentation using compositional model combining shape
  and appearance.
\newblock In {\em Proceedings of the IEEE conference on computer vision and
  pattern recognition}, pages 1788--1797, 2015.

\bibitem{wang2020mgcn}
Yiqun Wang, Jing Ren, Dong-Ming Yan, Jianwei Guo, Xiaopeng Zhang, and Peter
  Wonka.
\newblock Mgcn: descriptor learning using multiscale gcns.
\newblock {\em ACM Transactions on Graphics (TOG)}, 39(4):122--1, 2020.

\bibitem{wei2016dense}
Lingyu Wei, Qixing Huang, Duygu Ceylan, Etienne Vouga, and Hao Li.
\newblock Dense human body correspondences using convolutional networks.
\newblock In {\em Proceedings of the IEEE Conference on Computer Vision and
  Pattern Recognition}, pages 1544--1553, 2016.

\bibitem{xia2017joint}
Fangting Xia, Peng Wang, Xianjie Chen, and Alan~L Yuille.
\newblock Joint multi-person pose estimation and semantic part segmentation.
\newblock In {\em Proceedings of the IEEE conference on computer vision and
  pattern recognition}, pages 6769--6778, 2017.

\bibitem{xu2020rignet}
Zhan Xu, Yang Zhou, Evangelos Kalogerakis, Chris Landreth, and Karan Singh.
\newblock Rignet: neural rigging for articulated characters.
\newblock {\em ACM Transactions on Graphics (TOG)}, 39(4):58--1, 2020.

\bibitem{yu2017bodyfusion}
Tao Yu, Kaiwen Guo, Feng Xu, Yuan Dong, Zhaoqi Su, Jianhui Zhao, Jianguo Li,
  Qionghai Dai, and Yebin Liu.
\newblock Bodyfusion: Real-time capture of human motion and surface geometry
  using a single depth camera.
\newblock In {\em Proceedings of the IEEE International Conference on Computer
  Vision}, pages 910--919, 2017.

\bibitem{yu2018doublefusion}
Tao Yu, Zerong Zheng, Kaiwen Guo, Jianhui Zhao, Qionghai Dai, Hao Li, Gerard
  Pons-Moll, and Yebin Liu.
\newblock Doublefusion: Real-time capture of human performances with inner body
  shapes from a single depth sensor.
\newblock In {\em Proceedings of the IEEE conference on computer vision and
  pattern recognition}, pages 7287--7296, 2018.

\bibitem{zuffi2015stitched}
Silvia Zuffi and Michael~J Black.
\newblock The stitched puppet: A graphical model of 3d human shape and pose.
\newblock In {\em Proceedings of the IEEE Conference on Computer Vision and
  Pattern Recognition}, pages 3537--3546, 2015.

\bibitem{zuffi20173d}
Silvia Zuffi, Angjoo Kanazawa, David~W Jacobs, and Michael~J Black.
\newblock 3d menagerie: Modeling the 3d shape and pose of animals.
\newblock In {\em Proceedings of the IEEE conference on computer vision and
  pattern recognition}, pages 6365--6373, 2017.

\end{thebibliography}
}
\end{document}